\documentclass[12pt,letterpaper]{article}
\usepackage{amssymb,amsmath}
\usepackage{mathtools}
\usepackage{bm}
\usepackage{bbm}
\usepackage{graphicx}
\usepackage{enumerate}
\usepackage{caption}
\usepackage{url}
\usepackage{color}
\usepackage{placeins}
\usepackage{bm}
\usepackage{bbm}
\DeclareMathOperator*{\argmin}{\mbox{argmin}}
\DeclareMathOperator*{\argmax}{\mbox{argmax}}
\DeclareMathOperator*{\ave}{\mbox{ave}}
\DeclareMathOperator{\sign}{sign}

\newcommand{\sample}{X}
\newcommand{\BCl}{g_{\lambda}}

\newcommand{\YJl}{h_{\lambda}}
\newcommand{\YJlhat}{h_{\lambdahat}}
\newcommand{\rBCl}{\mathring{g}_{\lambda}}
\newcommand{\rYJl}{\mathring{h}_{\lambda}}
\newcommand{\eps}{\varepsilon}
\newcommand{\muhat}{\hat{\mu}}
\newcommand{\sigmahat}{\hat{\sigma}}
\newcommand{\lambdahat}{\hat{\lambda}}
\newcommand{\rhobw}{\rho_{\mbox{\small{bw}}}}
\newcommand{\SD}{\mbox{SD}}
\newcommand{\OD}{\mbox{OD}}
\newcommand{\bx}{\bf{x}}

\begin{document}

\def\spacingset#1{\renewcommand{\baselinestretch}%
{#1}\small\normalsize} \spacingset{1}

\title{\bf Transforming variables to central normality}		
\author{Jakob Raymaekers and Peter J. 
	  Rousseeuw \hspace{.1cm} \\
		Section of Statistics and Data Science, KU Leuven, 
		Belgium}
\date{November 21, 2020}
% Department of Mathematics,
\maketitle

\begin{abstract}
Many real data sets contain numerical features (variables) 
whose distribution is far from normal (gaussian).
Instead, their distribution is often skewed. 
In order to handle such data it is customary
to preprocess the variables to make them more normal.
The Box-Cox and Yeo-Johnson transformations are 
well-known tools for this.
However, the standard maximum likelihood estimator
of their transformation parameter is highly sensitive
to outliers, and will often try to move outliers
inward at the expense of the normality of the central 
part of the data.
We propose a modification of these 
transformations as well as an estimator of the 
transformation parameter that is robust to outliers, 
so the transformed data can be approximately normal
in the center and a few outliers may deviate from it.
It compares favorably to existing techniques in
an extensive simulation study and on real data.
\end{abstract}


%\spacingset{1.45}
\spacingset{1.1}

%\keywords{Anomaly detection \and Data preprocessing
 %\and Feature transformation
 %\and Outliers \and Symmetrization.}

%%%%%%%%%%%%%%%%%%%%%%%%%%%%%%%%%%%%%%%%%%%%%%%%%%%%%%%
\section{Introduction}
\label{sec:intro}

In machine learning and statistics, some numerical 
data features may be very nonnormal 
(nongaussian) and asymmetric (skewed) which often 
complicates the next steps of the analysis.
Therefore it is customary to preprocess
the data by transforming such features in order 
to bring them closer to normality, after
which it typically becomes easier to fit a
model or to make predictions. 
To be useful in practice, it must be possible to
automate this preprocessing step.

\begin{figure}
\includegraphics[width = 0.5\textwidth]
  {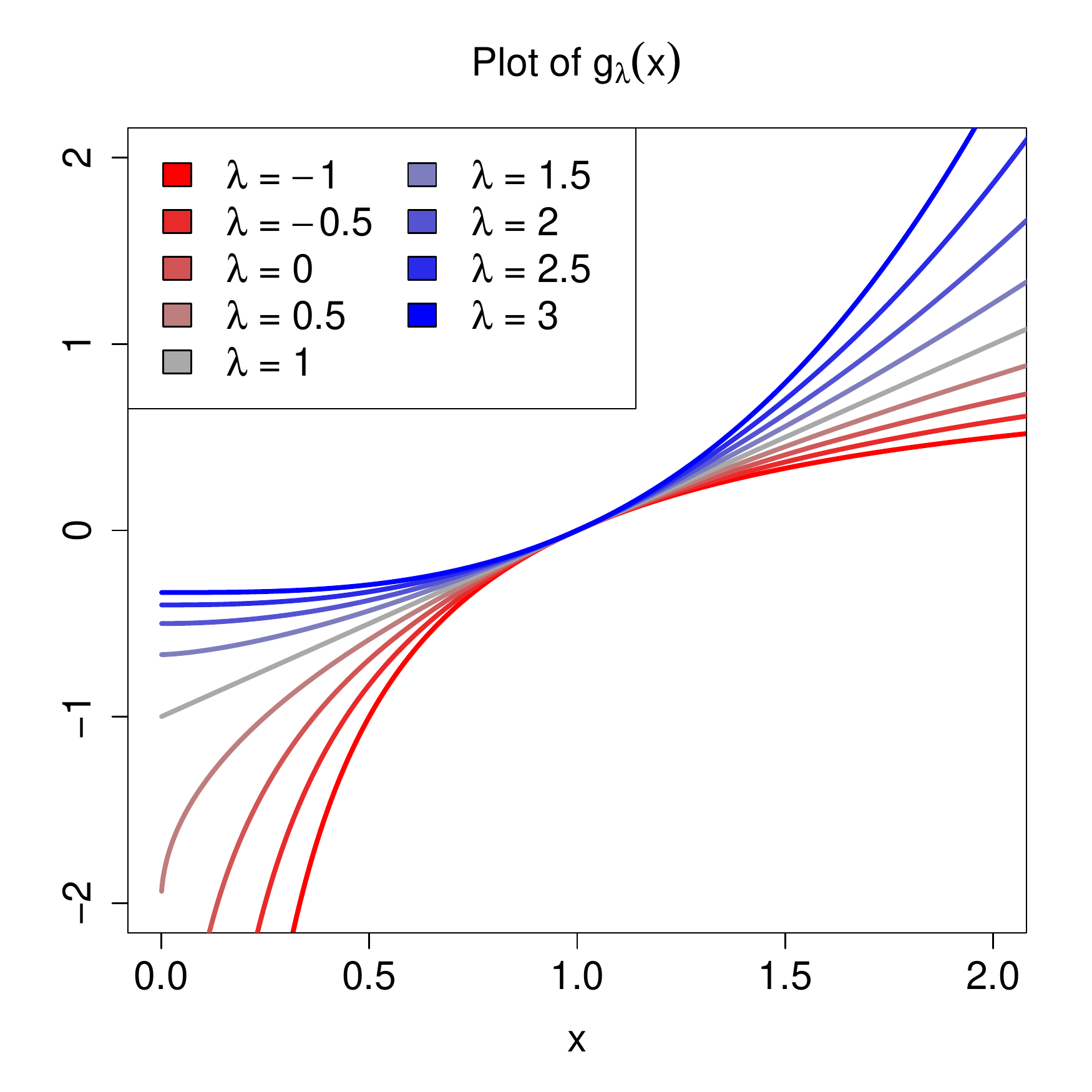}
\includegraphics[width = 0.5\textwidth]
  {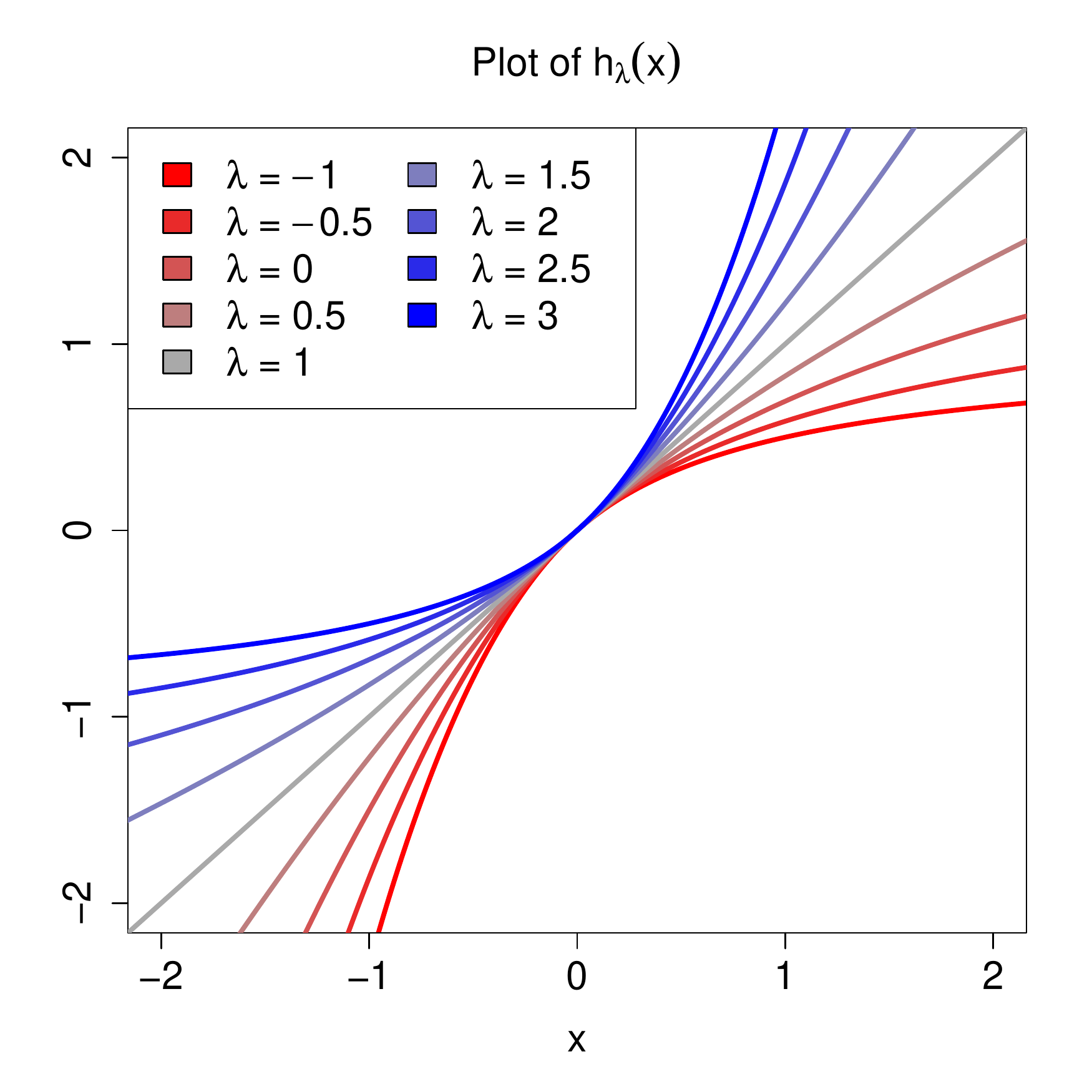}
\caption{The Box-Cox (left) and Yeo-Johnson (right) 
  transformations for several parameters $\lambda$.}
\label{fig:transformations}
\end{figure}

In order to transform a positive variable to give it
a more normal distribution one often resorts to
a power transformation (see e.g. \cite{Tukey1957}).
The most often used function is the Box-Cox (BC) 
power transform $\BCl$ studied by \cite{Box1964},
given by
\begin{equation}
\BCl(x) = 
\begin{cases}
(x^{\lambda} - 1) / \lambda &\mbox{ if } \lambda \neq 0\\
\log(x) &\mbox{ if } \lambda = 0.
\end{cases}
\end{equation}
Here $x$ stands for the observed feature, 
which is transformed to $\BCl(x)$ using a 
parameter $\lambda$.
A limitation of the family of BC transformations is 
that they are only applicable to positive data. 
To remedy this, Yeo and Johnson \cite{Yeo2000} proposed 
an alternative family of transformations that can deal 
with both positive and negative data. 
These Yeo-Johnson (YJ) transformations
$\YJl$ are given by
\begin{equation}
\YJl(x) = 
\begin{cases}
((1+x)^{\lambda} - 1) / \lambda &\mbox{ if } \lambda 
    \neq 0 \mbox{ and } x \geq 0\\
\log(1+x) &\mbox{ if } \lambda = 0 \mbox{ and } x \geq 0\\
-((1 - x)^{2 - \lambda} - 1)/(2 - \lambda) &\mbox{ if }
    \lambda \neq 2 \mbox{ and } x < 0\\
-\log(1 - x) &\mbox{ if } \lambda = 2 \mbox{ and } x < 0
\end{cases}
\end{equation}
and are also characterized by a parameter $\lambda$.
Figure \ref{fig:transformations} shows both of these 
transformations for a range of $\lambda$ values.
In both families $\lambda = 1$ yields a linear relation.
Transformations with $\lambda < 1$ compress the 
right tail of the distribution while expanding the left 
tail, making them suitable for transforming right-skewed 
distributions towards symmetry. 
Similarly, transformations with $\lambda > 1$ are 
designed to make left-skewed distributions more 
symmetrical.

Estimating the parameter $\lambda$ for the BC 
or YJ transformation is typically done using maximum 
likelihood, under the assumption that the transformed 
variable follows a normal distribution. 
However, it is well known that maximum likelihood 
estimation is very sensitive to outliers in the data, 
to the extent that a single outlier can have an 
arbitrarily large effect on the estimate. 
In the setting of transformation to normality, 
outliers can yield transformations for which the bulk 
of the transformed data follows a very skewed 
distribution, so no normality is attained at all.
In situations with outliers one would prefer to
make the non-outliers approximately normally
distributed, while the outliers may stay outlying.
So, our goal is to achieve {\it central normality},
where the transformed data look roughly normal in 
the center and a few outliers may deviate from it.
Fitting such a transformation is not easy, because
a point that is outlying 
in the original data may not be 
outlying in the transformed data, and vice versa.
The problem is that we do not know beforehand which 
points may turn out to be outliers in the 
optimally transformed data.

Some proposals exist in the literature to make 
the estimation of the parameter $\lambda$ in BC more 
robust against outliers, mainly in the context of 
transforming the response variable in a regression 
(\cite{Carroll1980,Marazzi2009,Riani2008}), but
here we are not in that setting. 
For the YJ transformation very few robust methods 
are available. In \cite{VdV2010} a trimmed maximum 
likelihood approach was explored, in which the 
objective is a trimmed sum of log likelihoods in
which the lowest terms are discarded. We will
study this method in more detail later.

Note that both the BC and YJ transformations 
suffer from the complication that their range depends 
on the parameter $\lambda$. In particular, for the BC 
transformation we have
\begin{equation} \label{eq:rangeBC}
\BCl(\mathbb{R}_0^+) = 
\begin{cases}
(-1/|\lambda|, \infty) & \mbox{ if }  \lambda > 0\\
\mathbb{R} & \mbox{ if }  \lambda = 0\\
(-\infty, 1/|\lambda|) & \mbox{ if }  \lambda < 0
\end{cases}
\end{equation}
whereas for the YJ transformation we have
\begin{equation} \label{eq:rangeYJ}
\YJl(\mathbb{R}) = 
\begin{cases}
(-1/|\lambda - 2|, \infty) & \mbox{ if } \lambda > 2\\
\mathbb{R} & \mbox{ if }  0 \leq \lambda \leq 2\\
(-\infty, 1/|\lambda|) & \mbox{ if }  \lambda < 0.
\end{cases}
\end{equation}
So, for certain values of $\lambda$ the range of the 
transformation is a half line.
This is not without consequences. 
First, most well-known symmetric distributions are 
supported on the entire line, so a perfect match
is impossible. 
More importantly, we argue that this can 
make outlier detection more difficult. 
Consider for instance the BC transformation with 
$\lambda = -1$ which has the range 
$g_{-1}(\mathbb{R}_0^+) = (-\infty, 1)$. 
Suppose we transform a data set $(x_1, \ldots, x_n)$ 
to $(g_{-1}(x_1), \ldots, g_{-1}(x_n))$. 
If we let $x_n \to \infty$ making it an extremely 
clear outlier in the original space, then 
$g_{-1}(x_n) \to 1$ in the transformed space. 
So a transformed outlier can be much closer to the bulk 
of the transformed data than the original outlier was 
in the original data. 
This is undesirable, since the outlier will be much 
harder to detect this way. 
This effect is magnified if $\lambda$ is estimated by
maximum likelihood, since this estimator will try to 
accommodate all observations, including the outliers.

\begin{figure}[!ht]
\includegraphics[width=0.5\textwidth]
  {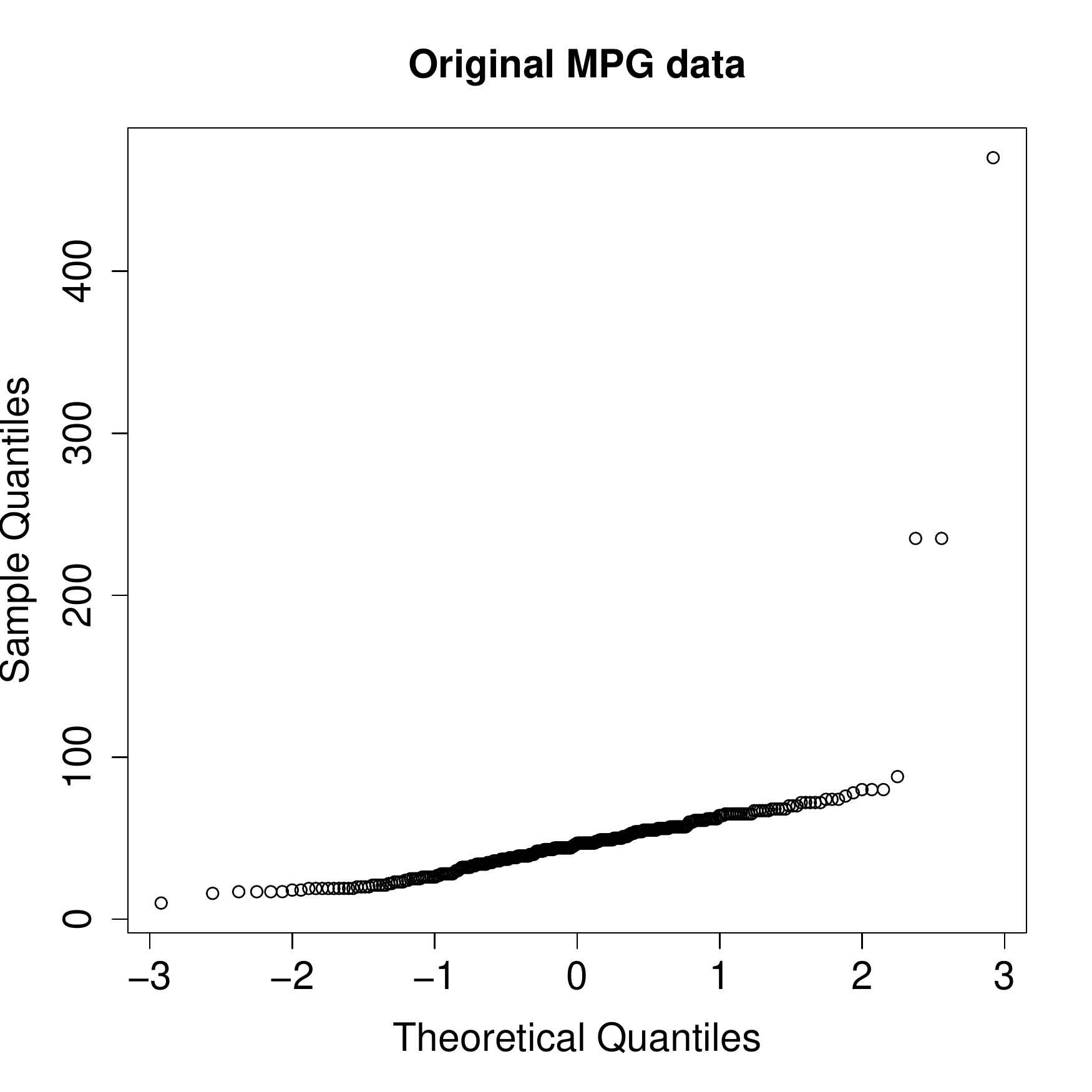}
\includegraphics[width=0.5\textwidth]
  {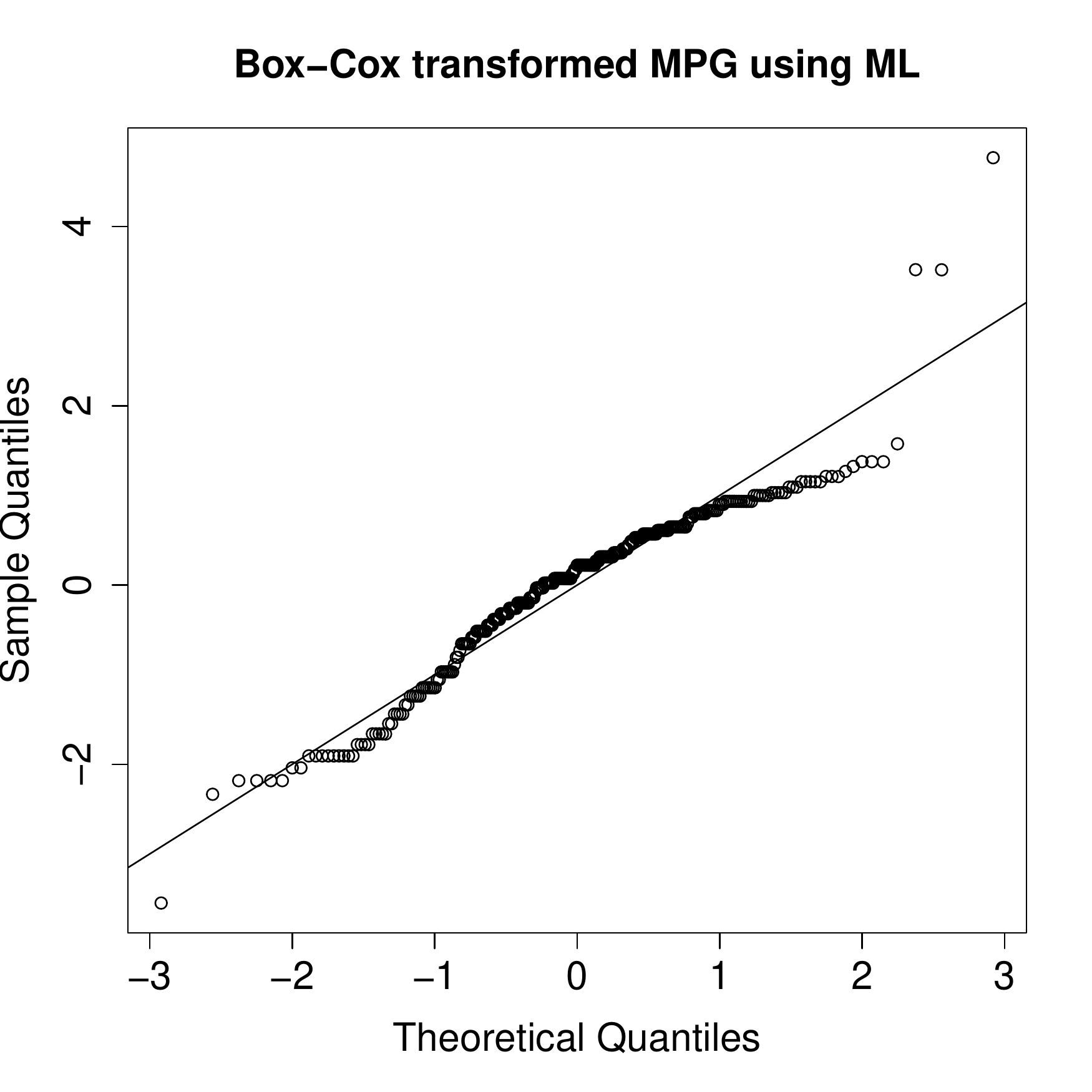}
\caption{Normal QQ-plot of the variable \texttt{MPG} in 
the Top Gear dataset (left) and the Box-Cox transformed 
variable using the maximum likelihood estimate of 
$\lambda$ (right). The ML estimate is heavily affected 
by the three outliers at the top, causing it to create 
skewness in the central part of the transformed data.}
\label{Fig:MPGQQ}
\end{figure}

We illustrate this point using the TopGear dataset 
\cite{Alfons2019} which contains information on 297 
cars, scraped from the website of the British 
television show Top Gear. 
We fit a Box-Cox transformation to the 
variable \texttt{miles per gallon} (\texttt{MPG}) 
which is strictly positive.
The left panel of Figure \ref{Fig:MPGQQ} shows the 
normal QQ-plot of the \texttt{MPG} variable before 
transformation. 
(That is, the horizontal axis contains as many
quantiles from the standard normal distribution as
there are sorted data values on the vertical axis.)
In this plot the majority of the observations seem to 
roughly follow a normal distribution, that is, many 
points in the QQ-plot lie close to a straight line. 
There are also three far outliers at the top, which 
correspond to the Chevrolet Volt and Vauxhall Ampera 
(both with 235 MPG) and the BMW i3 (with 470 MPG).
These cars are unusual because they derive most
of their power from a plug-in electric battery,
whereas the majority of the cars in the data set are
gas-powered.
The right panel of Figure \ref{Fig:MPGQQ} shows the 
Box-Cox transformed data using the maximum likelihood 
(ML) estimate $\hat{\lambda} = -0.11$, indicating that 
the BC transformation is fairly close to the log 
transform. 
We see that this transformation does not improve the 
normality of the MPG variable. 
Instead it tries to bring the three outliers into the 
fold, at the expense of causing skewness in the central 
part of the transformed data and creating an 
artificial outlier at the bottom.

\begin{figure}[!ht]
\includegraphics[width = 0.5 \textwidth]
  {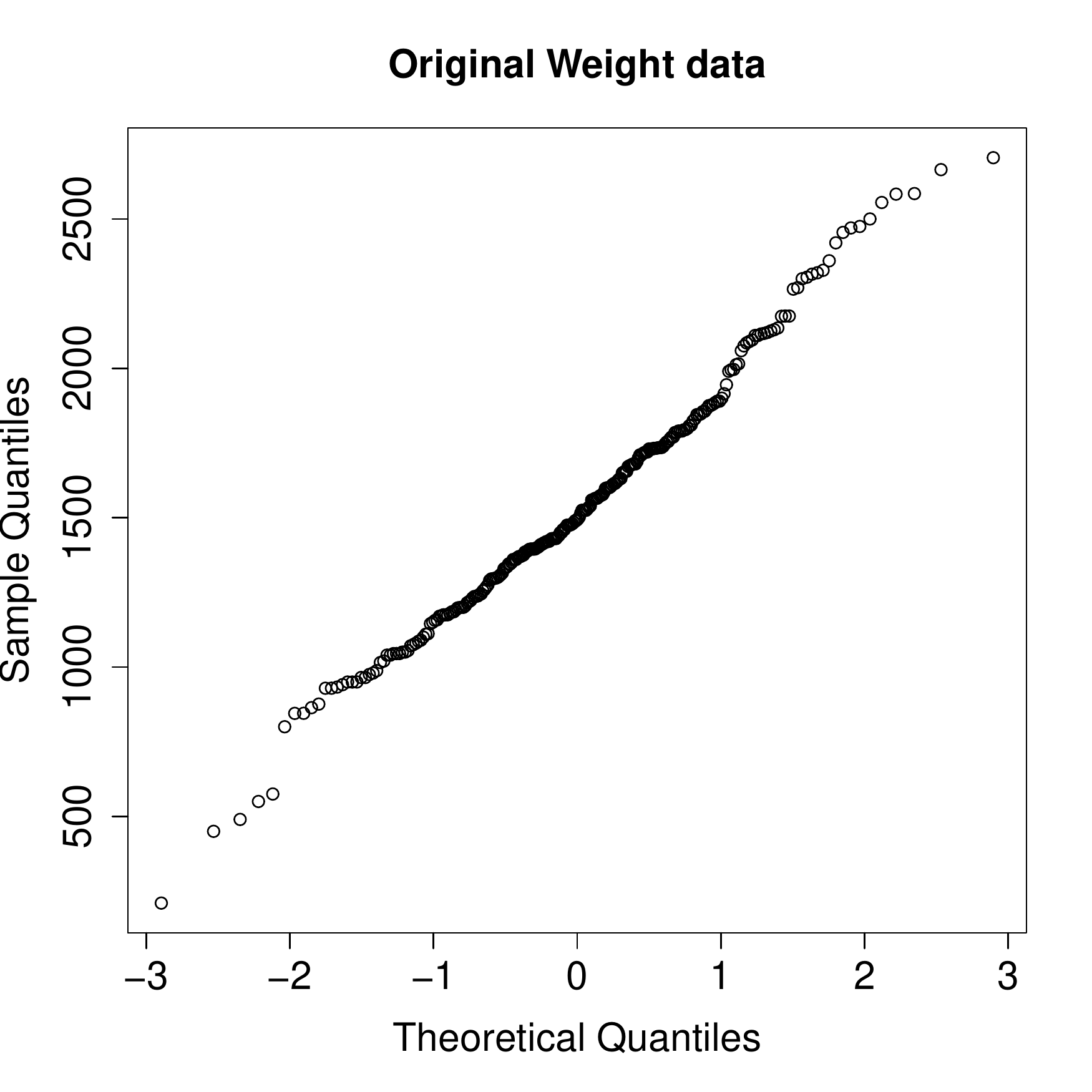}
\includegraphics[width = 0.5 \textwidth]
  {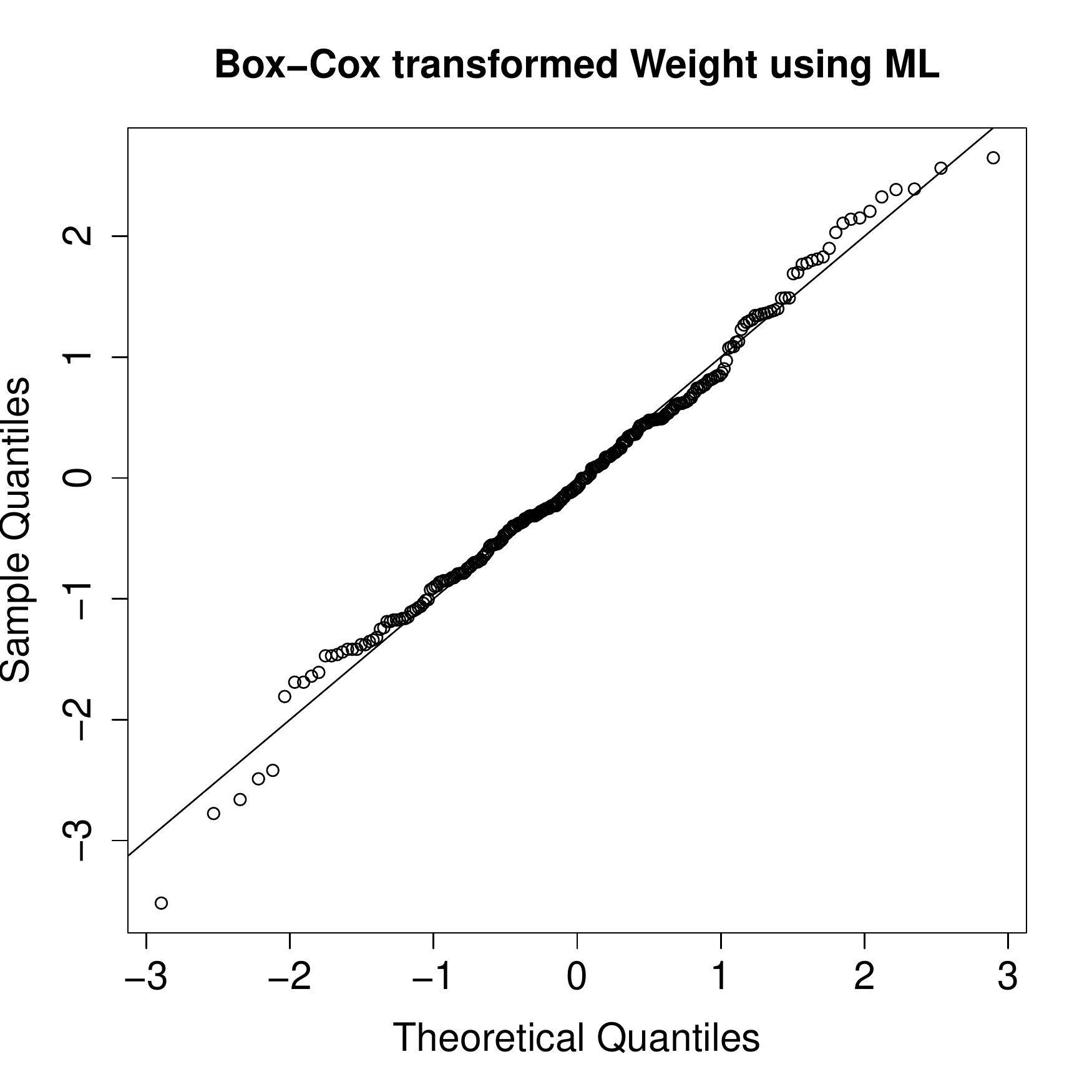}
\caption{Normal QQ-plot of the variable \texttt{Weight} 
in the Top Gear dataset (left) and the transformed 
variable using the ML estimate of $\lambda$ (right).
The transformation does not make the five outliers at 
the bottom stand out.}
\label{Fig:WeightQQ}
\end{figure}

The variable \texttt{Weight} shown in Figure 
\ref{Fig:WeightQQ} illustrates a different effect.
The original variable has one extreme and 4 
intermediate outliers at the bottom.
The extreme outlier is the Peugeot 107, whose
weight was erroneously listed as 210 kilograms,
and the next outlier is the tiny Renault Twizy
(410 kilograms).
Unlike the MPG variable the central part of these 
data is not very normal, as those points in the 
QQ-plot do not line up so well.
A transform that would make the central part more
straight would expose the outliers at the bottom more.
But instead the ML estimate is $\hat{\lambda} = 0.83$ 
hence close to $\lambda = 1$ which would correspond 
to not transforming the variable at all.
Whereas the \texttt{MPG} variable should not be 
transformed much but is, the \texttt{Weight} variable 
should be transformed but almost isn't.

In section \ref{sec:meth} we propose a new robust 
estimator for the parameter $\lambda$, and
compare its sensitivity curve to those of other
methods. 
Section \ref{sec:sim} presents a simulation to study 
the performance of several estimators on clean 
and contaminated data. 
Section \ref{sec:real} illustrates the proposed method 
on real data examples, and section \ref{sec:conc} 
concludes.

%%%%%%%%%%%%%%%%%%%%%%%%%%%%%%%%%%%%%%%%%%%%%%%%%%%%%%
\section{Methodology}
\label{sec:meth}

\subsection{Fitting a transformation by 
minimizing a robust criterion}
\label{sec:proposal}

The most popular way of estimating the $\lambda$ of the 
BC and YJ transformations is to use maximum likelihood 
(ML) under the assumption that the transformed variable 
follows a normal distribution, as will briefly be
summarized in subsection \ref{sec:RewML}.
However, it is well known that ML estimation is very 
sensitive to outliers in the data and other deviations 
from the assumed model. 
We therefore propose a different way of estimating the 
transformation parameter of a transformation.
 
Consider an ordered sample of univariate 
observations $\sample = (x_{(1)}, \ldots, x_{(n)})$ 
and a symmetric target distribution $F$. 
Suppose we want to estimate the parameter $\lambda$ of 
a nonlinear function $g_{\lambda}$ such that 
$g_{\lambda}(x_{(1)}), \ldots, g_{\lambda}(x_{(n)})$ 
come close to quantiles from the standard normal
cumulative distribution function $\Phi$. 
We propose to estimate $\lambda$ as:
\begin{equation}\label{eq:obj}
\lambdahat = \argmin_{\lambda} \; \sum_{i=1}^{n}{\rho 
   \left(\frac{g_{\lambda}(x_{(i)})-
	 \muhat_{\mbox{\tiny{M}}}}
	 {\sigmahat_{\mbox{\tiny{M}}}}
	 - \Phi^{-1}(p_i) \right)}\;.
\end{equation}
Here $\muhat_{\mbox{\tiny{M}}}$ 
is the Huber M-estimate of location
of the $g_{\lambda}(x_{(i)})$, and 
$\sigmahat_{\mbox{\tiny{M}}}$ is
their Huber M-estimate of scale.
Both are standard robust univariate estimators
(see \cite{Huber:RobStat}).
The $p_i = (i-1/3)/(n+1/3)$ are the usual equispaced 
probabilities that also yield the quantiles in the
QQ-plot (see, e.g., page 225 in \cite{Ureda}).
The function $\rho$ needs to be positive, even and 
continuously differentiable.
In least squares methods $\rho(t) = t^2$, but in our
situation there can be large absolute residuals 
$|\frac{g_{\lambda}(x_{(i)})-\muhat}{\sigmahat}
 - \Phi^{-1}(p_i)|$ 
caused by 
outlying values of $g_{\lambda}(x_{(i)})$.
In order to obtain a robust method we need a bounded 
$\rho$ function. 
We propose to use the well-known Tukey bisquare 
$\rho$-function given by
\begin{equation}\label{eq:rhobw}
 \rhobw(x) = 
\begin{cases}
  1 - (1 - (x/c)^2)^3 & \mbox{ if } |x| \leq c\\
  1 & \mbox{ if } |x| > c\,.
	\end{cases}
\end{equation}
The constant $c$ is a tuning parameter, which we set
to 0.5 by default here. See 
section A %\ref{secA:tuningc}
of the supplementary material for a motivation of 
this choice.

To calculate $\lambdahat$ numerically,
we use the \texttt{R} function \texttt{optimize()}
which relies on a combination of golden section
search and successive parabolic interpolation
to minimize the objective of \eqref{eq:obj}.

\subsection{Rectified Box-Cox and Yeo-Johnson 
            transformations}
\label{sec:rectif}

In this section we propose a modification of the 
classical BC and YJ transformations, called the 
\textit{rectified} BC and YJ transformations.
They make a continuously differentiable switch to 
a linear transformation in the tails of the 
BC and YJ functions.
The purpose of these modified transformations is 
to remedy two issues. 
First, the range of the classical BC and YJ
transformations depends on $\lambda$ and is often
only a half line.
And second, as argued in the introduction, the 
classical transformations often push outliers 
closer to the majority of the data, which makes 
the outliers harder to detect. 
Instead the range of the proposed modified 
transformations is always the entire real line, 
and it becomes less likely that outliers are masked 
by the transformation.

For $\lambda < 1$, the BC transformation is 
designed to make right-skewed distributions more 
symmetrical, and is bounded from above. 
In this case we define the rectified BC transformation 
as follows. Consider an upper constant $C_u > 1$. 
The rectified BC transformation $\rBCl$ is defined as
\begin{equation}\label{eq:BCright}
\rBCl(x) = 
\begin{cases}
\BCl(x) &\mbox{ if } x \leq C_u\\
\BCl(C_u) + (x - C_u) \BCl'(C_u) &
            \mbox{ if } x > C_u\,.
\end{cases}
\end{equation}
Similarly, for $\lambda > 1$ and a positive lower 
constant $C_\ell < 1$ we put 
\begin{equation}\label{eq:BCleft}
\rBCl(x) = 
\begin{cases}
\BCl(C_\ell) + (x - C_\ell) \BCl'(C_\ell) &
   \mbox{ if } x < C_\ell\\
\BCl(x) &\mbox{ if } x \geq C_\ell\,.
\end{cases}
\end{equation}

\begin{figure}
\includegraphics[width = 0.5\textwidth]
  {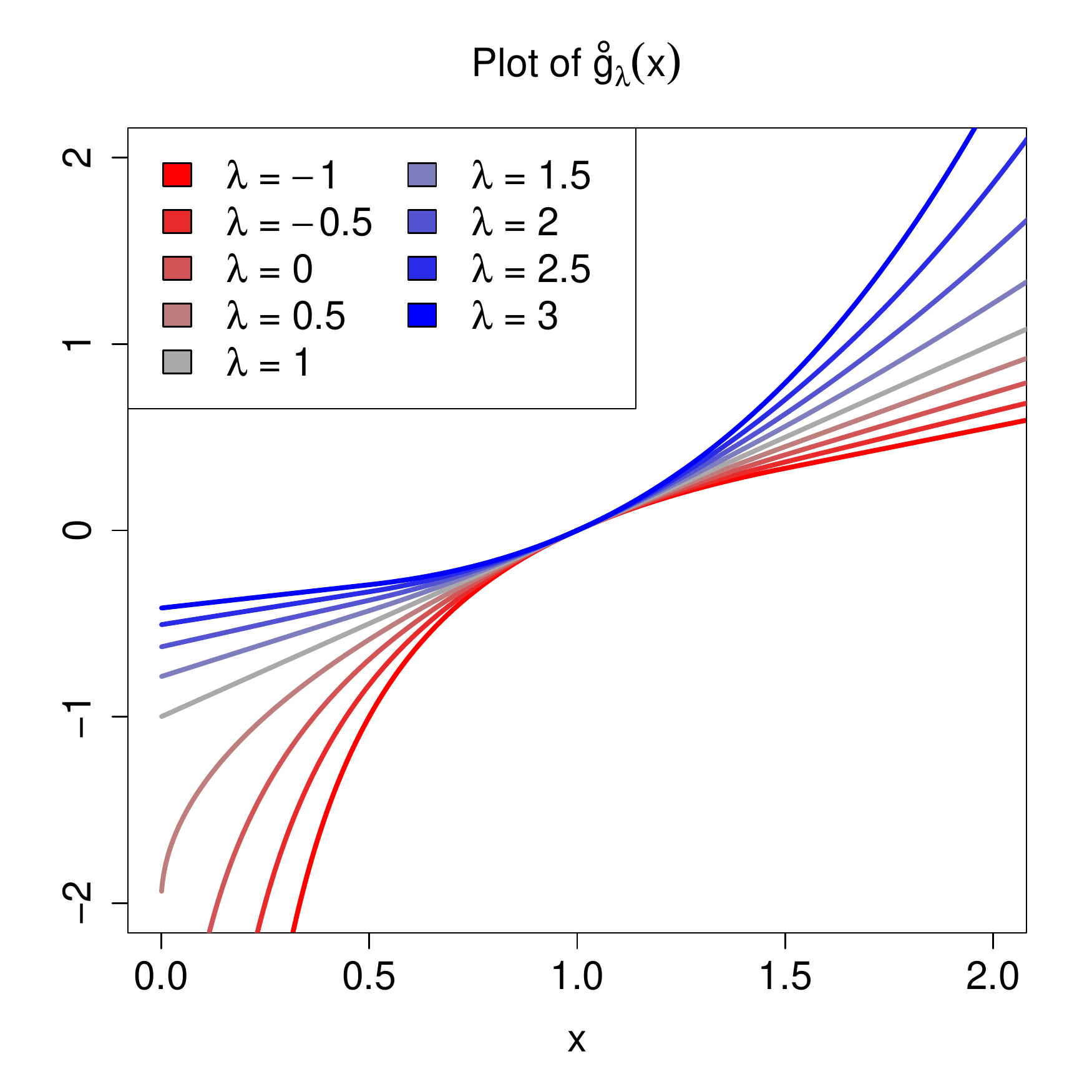}
\includegraphics[width = 0.5\textwidth]
  {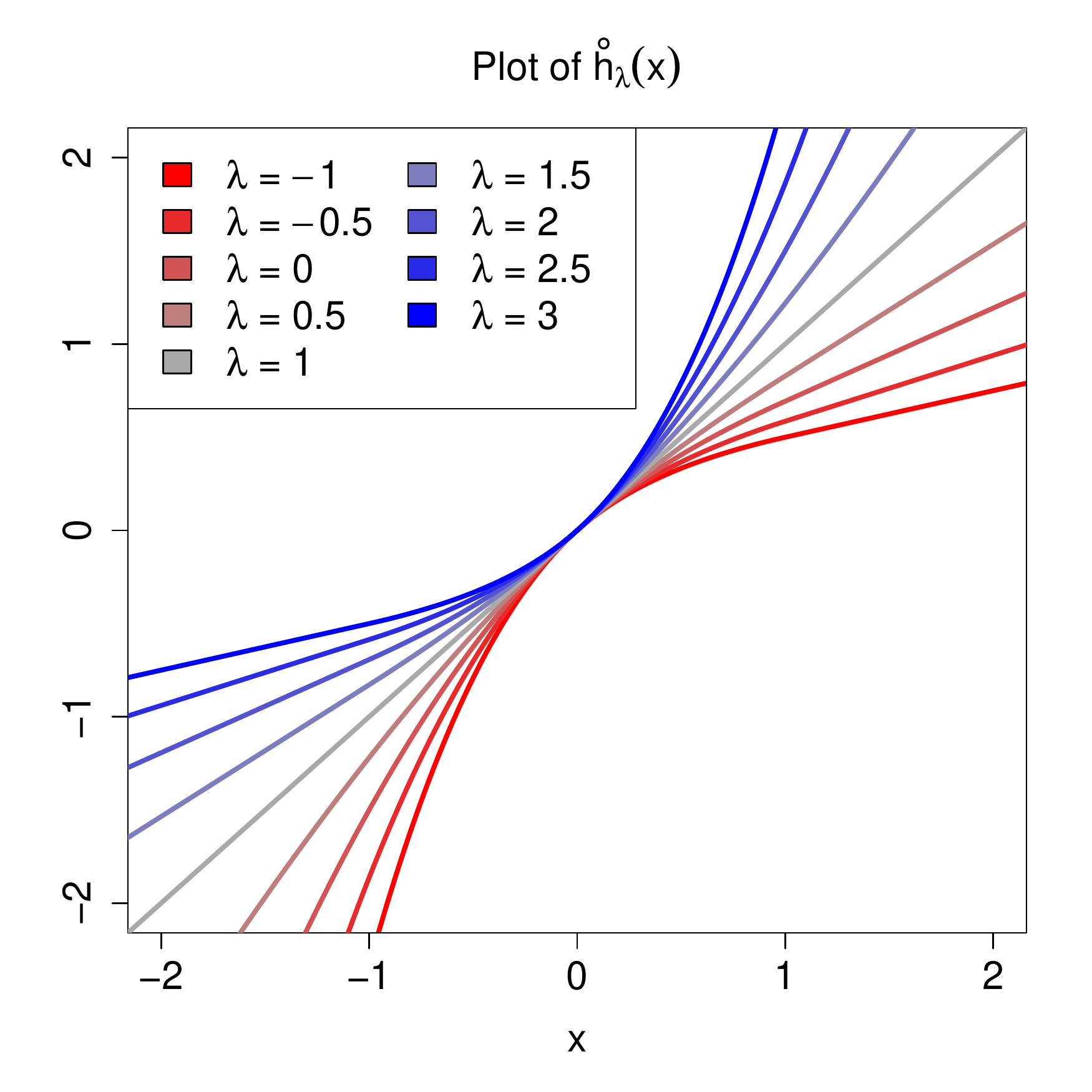}
\caption{The rectified Box-Cox (left) and Yeo-Johnson
(right) transformations for a range of parameters $\lambda$.
They look quite similar to the original transformations
in Figure \ref{fig:transformations} but contract less
on the right when $\lambda < 1$, and contract less on the 
left when $\lambda > 1$.}
\label{fig:adjtransformations}
\end{figure}

For the YJ transformation we construct rectified 
counterparts in a similar fashion. 
For $\lambda < 1$ and a value $C_u > 0$ we define the 
rectified YJ transformation $\rYJl(x)$ as in 
\eqref{eq:BCright} with $\BCl$ replaced by $\YJl$:
\begin{equation}\label{eq:YJright}
\rYJl(x) = 
\begin{cases}
\YJl(x) &\mbox{ if } x \leq C_u\\
\YJl(C_u) + (x - C_u) \YJl'(C_u) &
    \mbox{ if } x > C_u\,.
\end{cases}
\end{equation}
Analogously, for $\lambda > 1$ and $C_\ell < 0$
we define $\rYJl(x)$ as in \eqref{eq:BCleft}: 
\begin{equation}\label{eq:YJleft}
\rYJl(x) = 
\begin{cases}
\YJl(C_\ell) + (x - C_\ell) \YJl'(C_\ell) &
    \mbox{ if } x < C_\ell\\
\YJl(x) &\mbox{ if } x \geq C_\ell\,.
\end{cases}
\end{equation}
Figure \ref{fig:adjtransformations} shows 
such rectified BC and YJ transformations.\\ 

What are good choices of $C_\ell$ and $C_u$? 
Since the original data is often asymmetric,
we cannot just use a center (like the median) plus or
minus a fixed number of (robust) standard deviations. 
Instead we set $C_\ell$ equal to the first quartile of the 
original data, and for $C_u$ we take the third quartile. 
Other choices could be used, but more extreme quantiles 
would yield a higher sensitivity to outliers.

%%%%%%%%%%%%%%%%%%%%%%%%%%%%%%%%%%%%%%%%%%
\subsection{Reweighted maximum likelihood} 
\label{sec:RewML}

We now describe a reweighting scheme to increase 
the accuracy of the estimated $\lambdahat$ while 
preserving its robustness.
For a data set $x_1, \ldots, x_n$ the classical maximum 
likelihood estimator for the Yeo-Johnson transformation 
parameter $\lambda$ is given by the $\lambda$ which 
maximizes the normal loglikelihood. 
After removing constant terms this can be written as:
\begin{equation}
\label{eq:ML}
  \lambdahat_{\mbox{\tiny{ML}}}^
	{\mbox{\tiny{YJ}}} =
	\argmax_{\lambda}\;
  {\sum_{i = 1}^{n}{ -\frac{1}{2} 
	\log(\sigmahat_{\mbox{\tiny{ML}},\lambda}^2) + 
	(\lambda - 1) \sign(x_i) \log(|x_i|+1)}}
\end{equation}
where $\sigmahat_{\mbox{\tiny{ML}},\lambda}^2$
is the maximum likelihood scale of the transformed 
data given by
\begin{equation}
\label{eq:musigmaML}
  \sigmahat_{\mbox{\tiny{ML}},\lambda}^2
	= \frac{1}{n} \sum_{i = 1}^{n}{(\YJl(x_i) - 
	   \muhat_{\mbox{\tiny{ML}},\lambda})^2}
		\quad \mbox{where} \quad
	\muhat_{\mbox{\tiny{ML}},\lambda}
	= \frac{1}{n} \sum_{i = 1}^{n}{\YJl(x_i)} \;.
\end{equation}
The last term in \eqref{eq:ML} comes from the 
derivative of the YJ transformation.
Criterion \eqref{eq:ML} is sensitive to outliers
since it depends on a classical variance and the 
unbounded term $\log(1+|x_i|)$. 
This can be remedied by using weights. 
Given a set of weights $W = (w_1,\ldots, w_n)$ we 
define a weighted maximum likelihood (WML) 
estimator by
\begin{equation}
\label{eq:WMLYJ}
  \lambdahat_{\mbox{\tiny{WML}}}^
	  {\mbox{\tiny{YJ}}} = 
  \argmax_{\lambda}\;{\sum_{i = 1}^{n}{w_i 
  \left[ -\frac{1}{2} \log(
	\sigmahat_{\mbox{\tiny{W}},\lambda}^2) 
   + (\lambda - 1) \sign(x_i) \log(1+|x_i|)\right]}}
\end{equation}
where $\sigmahat_{\mbox{\tiny{W}},\lambda}^2$ now 
denotes the weighted variance of the transformed 
data:
\begin{equation}
\label{eq:musigmaW}
  \sigmahat_{\mbox{\tiny{W}},\lambda}^2
	= \frac{\sum_{i = 1}^{n}{w_i (\YJl(x_i) - 
	   \muhat_{\mbox{\tiny{W}},\lambda})^2}}
     {\sum_{i = 1}^{n}{w_i}}
		\quad \mbox{where} \quad
	\muhat_{\mbox{\tiny{W}},\lambda}
	=  \frac{\sum_{i = 1}^{n}{w_i\, \YJl(x_i)}}
     {\sum_{i = 1}^{n}{w_i}} \;.
\end{equation}
If the weights appropriately downweight the outliers 
in the data, the WML criterion yields a more robust 
estimate of the transformation parameter.

For the BC transform the reasoning is analogous, the 
only change being the final term that comes
from the derivative of the BC transform. This yields
\begin{equation}
\label{eq:WMLBC}
  \lambdahat_{\mbox{\tiny{WML}}}^
	  {\mbox{\tiny{BC}}} = 
  \argmax_{\lambda}\;{\sum_{i = 1}^{n}{w_i 
  \left[ -\frac{1}{2} \log(
	\sigmahat_{\mbox{\tiny{W}},\lambda}^2) 
   + (\lambda - 1) \log(x_i)\right]}}\;\;.
\end{equation}

In general, finding robust data weights is 
not an easy task. 
The problem is that the observed data 
$\sample = (x_1, \ldots, x_n)$ can have a (very) 
skewed distribution and there is no straightforward
way to know which points will be outliers in the
transformed data when $\lambda$ is unknown.
But suppose that we have a rough initial estimate 
$\lambda_0$ of $\lambda$. 
We can then transform the data with $\lambda_0$ 
yielding $h_{\lambda_0}(\sample) = 
(h_{\lambda_0}(x_1), \ldots, h_{\lambda_0}(x_n))$, which 
should be a lot more symmetric than the original data.
We can therefore compute weights on 
$h_{\lambda_0}(\sample)$ using a classical weight 
function. Here we will use the "hard rejection rule" 
given by
\begin{equation} \label{eq:weights}
  w_i = 
  \begin{cases}
   1 &\mbox{ if } |h_{\lambda_0}(x_i)
	    -\muhat| \leqslant \Phi^{-1}(0.995)\,\sigmahat\\
	 0 &\mbox{ if } |h_{\lambda_0}(x_i)
	    -\muhat| > \Phi^{-1}(0.995)\,\sigmahat\\
	\end{cases}	
	\end{equation}
with $\muhat$ and $\sigmahat$ as in \eqref{eq:obj}.
With these weights we can compute a reweighted 
estimate $\lambdahat_1$ by the WML estimator in 
\eqref{eq:WMLYJ}. 
Of course, the robustness of the reweighted estimator 
will depend strongly on the robustness of the initial 
estimate $\lambda_0$\,.

Note that the reweighting step can be iterated,
yielding a multistep weighted ML estimator. 
In simulation studies we found that more than 2 
reweighting steps provided no further improvement in 
terms of accuracy (these results are not shown for 
brevity). We will always use two reweighting 
steps from here onward.
 
%%%%%%%%%%%%%%%%%%%%%%%%%%%%%%%%%%%%%%%%%%
\subsection{The proposed method}
\label{sec:prop}

Combining the above ideas, our proposed 
reweighted maximum likelihood (RewML) method
consists of the following steps:
\begin{itemize}
\item Step 1. Compute the initial estimate 
  $\lambda_0$ by maximizing the robust 
	criterion~\eqref{eq:obj}. 
	When fitting a Box-Cox transformation, 
	plug in the rectified function $\rBCl$. 
	When fitting a Yeo-Johnson transformation, use
	the rectified function $\rYJl$. Note that the
	rectified transforms are only used in this first
	step.
\item Step 2. Using $\lambda_0$ as 
  starting value, 
  compute the reweighted ML estimate from 
	\eqref{eq:WMLBC} when fitting the unrectified 
	Box-Cox transform $\BCl$\,, and from 
	\eqref{eq:WMLYJ} when	fitting the unrectified
	Yeo-Johnson transform $\YJl$\,.
\item Step 3. Repeat step 2 once and stop.
\end{itemize}

%%%%%%%%%%%%%%%%%%%%%%%%%%%%%%%%%%%%%%%%%%
\subsection{Other estimators of $\lambda$}
\label{sec:otherEst}

We will compare our proposal with two existing
robust methods.\\

The first is the robustified ML estimator 
proposed by Carroll in 1980 (\cite{Carroll1980}).
The idea was to replace the variance 
$\sigmahat_{\mbox{\tiny{ML}},\lambda}^2$
in the ML formula \eqref{eq:ML} by a robust variance 
estimate of the transformed data.
Carroll's method was proposed for the BC transformation, 
but the idea can be extended naturally to the estimation 
of the parameter of the YJ transformation. 
The estimator is then given by 
\begin{equation}
\label{eq:Carroll}
  \hat{\lambda}_{\mbox{Carroll}} =
	\argmax_{\lambda}\;{\sum_{i = 1}^{n}{ -\frac{1}{2}
	\log(\hat{\sigma}_{\text{M},\lambda}^2) + 
	(\lambda - 1) \sign(x_i) \log(1+|x_i|)}}
\end{equation}
where $\hat{\sigma}_{\text{M},\lambda}$ denotes the 
usual Huber M-estimate of scale \cite{Huber:RobStat} of 
the transformed data set $(\YJl(x_1), \ldots, \YJl(x_n))$.

The second method is the maximum trimmed likelihood 
(MTL) estimator of \cite{VdV2010}. 
Given a data set of size $n$, and a fixed number $h$
that has to satisfy $\lceil\frac{n}{2}\rceil<h < n$, 
this method looks for the parameter $\lambdahat$ which 
produces a subset of $h$ consecutive observations which 
maximize the ML criterion \eqref{eq:ML}.

%%%%%%%%%%%%%%%%%%%%%%%%%%%%%%%%%%%
\subsection{Sensitivity curves} 
\label{sec:SC}

In order to assess robustness against an outlier, 
stylized sensitivity curves were introduced on 
page 96 of \cite{Andrews1972}. 
For a given estimator $T$ and a cumulative distribution 
function $F$ they are constructed as follows:

\begin{enumerate}
\item Generate a stylized pseudo data set $\sample^0$
of size $n-1$:
\begin{equation*}
  \sample^0 = (x_1, \ldots, x_{n-1}) = 
	\left(F^{-1}(p_1), \ldots,
	F^{-1}(p_{n-1})\right)
\end{equation*}
where the $p_i$ for $i,\ldots,n-1$ are equispaced 
probabilities that are symmetric about 1/2. 
We can for instance use $p_i = i/n$.
\item Add to this stylized data set a variable point 
$z$ to obtain 
\begin{equation*}
  \sample_z = (x_1, \ldots, x_{n-1}, z).
\end{equation*}
\item Calculate the sensitivity curve in $z$ by 
\begin{equation*}
  \mbox{SC}_n(z) \coloneqq n\left(T(\sample_z) - 
	 T(\sample^0)\right)
\end{equation*}
where $z$ ranges over a grid chosen by the user.
The purpose of the factor $n$ is to put 
sensitivity curves with different values of $n$ on 
a similar scale.
\end{enumerate}

\begin{figure}
\centering
\includegraphics[width = 0.6\textwidth]
  {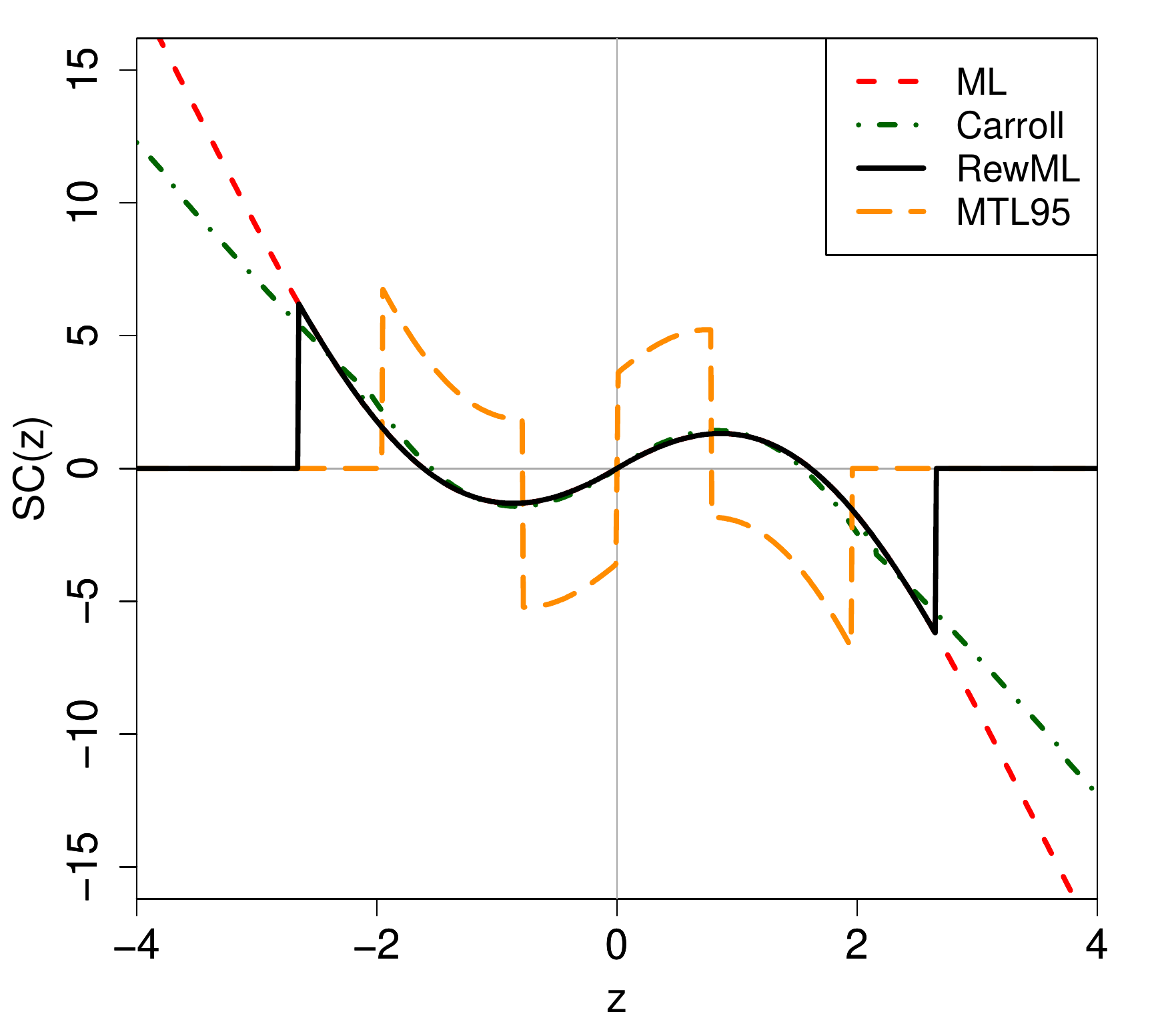}
\includegraphics[width = 0.6\textwidth]
  {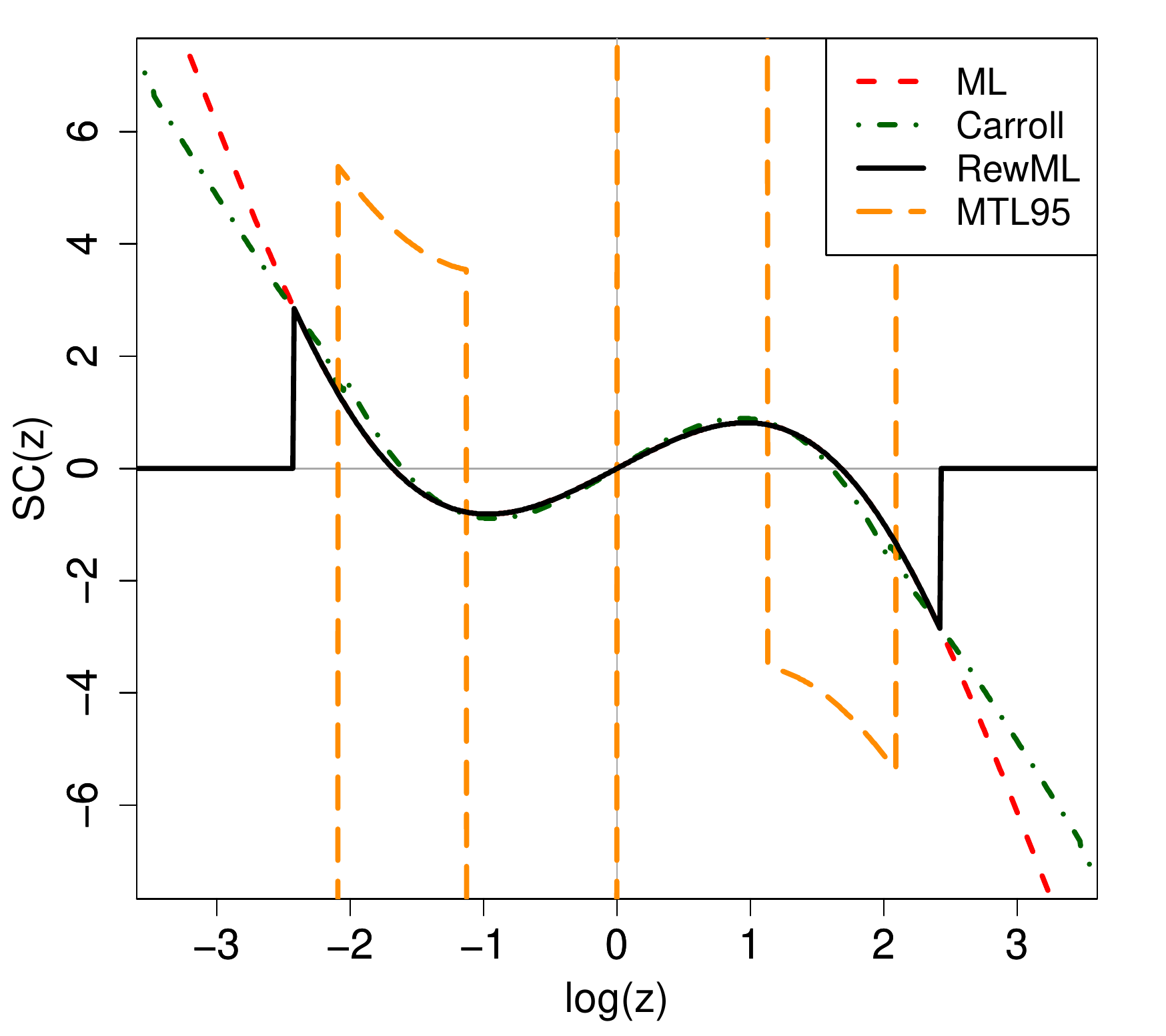}
\caption{Sensitivity curves of estimators of the 
  parameter $\lambda$ in the Yeo-Johnson 
	(top) and Box-Cox (bottom)
	transformations, with	sample size $n = 100$.}
\label{fig:SC}	
\end{figure}

The top panel of Figure \ref{fig:SC} shows the
sensitivity curves for several estimators of the
parameter $\lambda$ of the YJ transformation.
We chose $F = \Phi$ so the true transformation 
parameter $\lambda$ is 1, and $n=100$.
The maximum likelihood estimator ML of \eqref{eq:ML}
has an unbounded sensitivity curve, which is 
undesirable as it means that a single outlier can 
move $\lambdahat$ arbitrarily far away. 
The estimator of Carroll \eqref{eq:Carroll} has the 
same property, but is less affected in the sense that 
for a high $|z|$ the value of $|\mbox{SC}_n(z)|$ is
smaller than for the ML estimator. 
The RewML estimator that we proposed in 
subsection \ref{sec:prop} has a sensitivity curve
that lies close to that of the ML in the central region
of $z$, and becomes exactly zero for more extreme 
values of $|z|$.
Such a sensitivity curve is called {\it redescending},
meaning that it goes back to zero.
Therefore a far outlier has little effect on the 
resulting estimate. 
We also show MTL95, the trimmed likelihood estimator 
described in subsection \ref{sec:otherEst}
with $h/n = 95\%$. Its sensitivity curve is also 
redescending, but in the central region it is more
erratic with several jumps.

The lower panel of Figure \ref{fig:SC} shows the 
sensitivity curves for the Box-Cox transformation 
when the true parameter is $\lambda = 0$, i.e. the 
clean data follows a lognormal distribution $F$. 
We now put $\log(z)$ on the horizontal axis, since 
this makes the plot more comparable to that for
Yeo-Johnson in the top panel.
Also here the ML and Carroll's estimator have an 
unbounded sensitivity curve. Our RewML estimator 
has a redescending SC which again behaves similarly to 
the classical ML for small $|\log(z)|$, whereas the 
sensitivity to an extreme outlier is zero. 
The maximal trimmed likelihood estimator MTL95 
has large jumps reaching values over 40 in the 
central region.
Those peaks are not shown because the other curves 
would be hard to distinguish at that scale.\\

%%%%%%%%%%%%%%%%%%%%%%%%%%%%%%%%%%%%%%%%%%%%%%%%%%%%%
\section{Simulation}
\label{sec:sim}

%%%%%%%%%%%%%%%%%%%%%%%%%%%%%
\subsection{Compared Methods}
\label{sec:SimMethods}

For the Box-Cox as well as the Yeo-Johnson transformations
we perform a simulation study to compare the performance 
of several methods, including our proposal. 
The estimators under consideration are:
\begin{enumerate}
\item \textbf{ML}: the classical maximum likelihood 
  estimator given by \eqref{eq:ML}, or by 
	\eqref{eq:WMLBC} with all $w_i=1$.
\item \textbf{Carroll}: the robustified maximum likelihood 
  estimator of \cite{Carroll1980} given by 
	\eqref{eq:Carroll}.
\item \textbf{MTL}: the maximum trimmed likelihood 
  estimator of \cite{VdV2010}. The notation MTL90 stands
	for the version with $h/n = 90\%$\,.
\item \textbf{RewML}: the proposed reweighted maximum 
  likelihood estimator described in subsection 
	\ref{sec:prop}.
\item \textbf{RewMLnr}: a variation on RewML in which
  the first step of subsection \ref{sec:prop} applies
	\eqref{eq:obj} to the original Box-Cox or Yeo-Johnson
	transform instead of their rectified versions.
	This is not intended as a proposal, but included in
	order to show the advantage of rectification.
\end{enumerate}

%%%%%%%%%%%%%%%%%%%%%%%%%%%%
\subsection{Data generation}
\label{sec:generate}

We generate clean data sets as well as data 
with a fraction $\eps$ of outliers. 
The clean data are produced 
by generating a sample of size $n$ from the standard 
normal distribution, after which the inverse of the BC 
or YJ transformation with a given $\lambda$ is applied. 
For contaminated data we replace a percentage $\eps$
of the standard normal data by outliers at a fixed
position before the inverse transformation is applied.
For each such combination of $\lambda$ and $\eps$ we 
generate $m = 100$ data sets.\\

To be more precise, the percentage $\eps$ of 
contaminated points takes on the values 0, 0.05, 0.1, 
and 0.15, where $\eps=0$ corresponds to 
uncontaminated data.
For the YJ transformation we take the true 
transformation parameter $\lambda$ equal to
0.5, 1.0, or 1.5\,. We chose these values because
for $\lambda$ between 0 and 2 the range of YJ 
given by \eqref{eq:rangeYJ} is the entire real line, 
so the inverse of YJ is defined for all real numbers.
For the BC transformation we take $\lambda = 0$ 
for which the range \eqref{eq:rangeBC} is also the 
real line.
For a given combination of $\eps$ and $\lambda$
the steps of the data generation are:
\begin{enumerate}
\item Generate a sample $Y = (y_1, \ldots, y_n)$ from 
  the standard normal distribution. Let $k >0$ be
	a positive parameter. 
	Then replace a fraction $\eps$ of the points in $Y$ 
	by $k$ itself when $\lambda \leqslant 1$, and by 
	$-k$ when $\lambda>1$.
\item Apply the inverse BC transformation to $Y$,
  yielding the data set $\sample$ given by\linebreak  
	$\sample = (\BCl^{-1}(y_1), \ldots, \BCl^{-1}(y_n))$. 
	For YJ we put 
	$\sample = (\YJl^{-1}(y_1), \ldots, \YJl^{-1}(y_n))$.
\item Estimate $\lambda$ from $\sample$ using the 
  methods described in subsection \ref{sec:SimMethods}.
\end{enumerate}
The parameter $k$ characterizing the position of the
contamination is an integer that we let range
from 0 to 10.

We then estimate the bias and mean squared error (MSE) 
of each method by
\begin{align*}
  \mbox{bias} &:= \ave_{j=1}^{n}{
               (\hat{\lambda}_j - \lambda)}\\
  \mbox{MSE}  &:= \ave_{j=1}^{n}{
               (\hat{\lambda}_j - \lambda)^2}
\end{align*}
where $j = 1,\ldots,m$ ranges over the
generated data sets.

%%%%%%%%%%%%%%%%%%%%%%%%%%%%%%%%%%%%%%%%%%%%%%%%%%%%%%%
\subsection{Results for the Yeo-Johnson transformation}
\label{sec:ResultsYJ}

We first consider the effect of an increasing 
percentage of contamination on the different estimators. 
In this setting we fix the position of the contamination 
by setting $k=10$.
(The results for $k=6$ are qualitatively similar, 
as can be seen in section B %\ref{secA:incc} 
of the supplementary material.)
Figure \ref{fig:allepsplotsYJ} shows the bias and MSE 
of the estimators for an increasing contamination
percentage $\eps$ on the horizontal axis. 
The results in the top row are for data generated with
$\lambda=0.5$, whereas the middle row was generated
with $\lambda=1$ and the bottom row with 
$\lambda=1.5$\,.
In all rows the classical ML and the Carroll estimator 
have the largest bias and MSE, meaning they react 
strongly to far outliers, as suggested by their 
unbounded sensitivity curves in Figure \ref{fig:SC}.
In contrast with this both RewML and RewMLnr perform much 
better as their bias and MSE are closer to zero.
Up to about $5\%$ of outliers their curves are almost
indistinguishable, but beyond that RewML outperforms
RewMLnr by a widening margin. This indicates that using
the rectified YJ transform in the first step of the
estimator (see subsection \ref{sec:prop}) is more robust
than using the plain YJ in that step, even though 
the goal of the entire 3-step procedure RewML is to 
estimate the $\lambda$ of the plain YJ transform.

\begin{figure}[!ht]
\begin{centering}
\includegraphics[width = 0.83\textwidth]
  {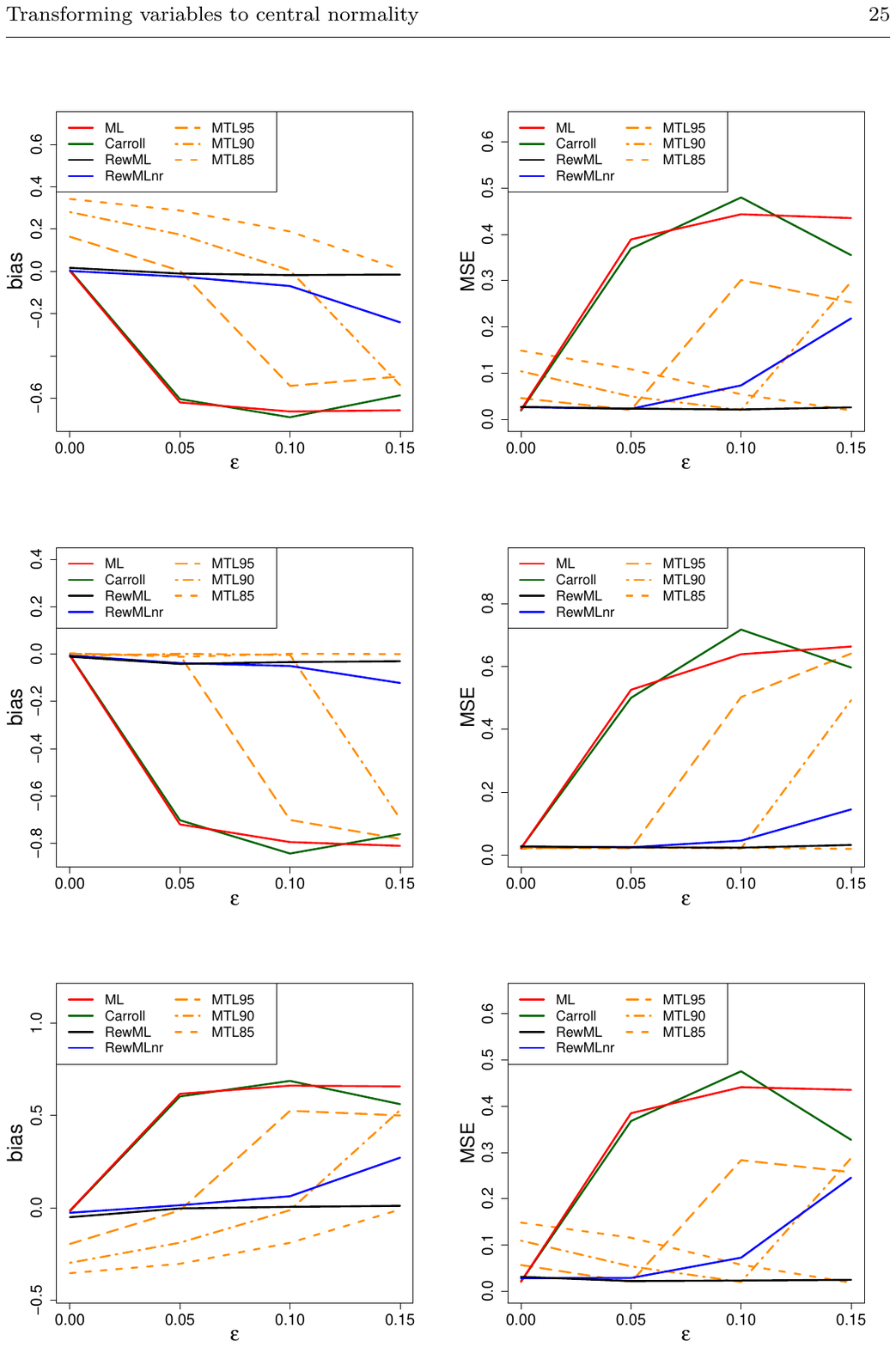}\\
\end{centering}
\caption{Bias (left) and MSE (right) of the
  estimated $\lambdahat$ of the 
  Yeo-Johnson transformation as a function of the 
	percentage $\eps$ of outliers, when the location 
	of the outliers is determined by setting $k=10$.
  The true parameter $\lambda$ used to generate
	the data is 0.5 in the top row, 1.0 in the 
	middle row, and 1.5 in the bottom row.}
\label{fig:allepsplotsYJ}
\end{figure}

In the same Figure \ref{fig:allepsplotsYJ} we see 
the behavior of the maximum trimmed likelihood 
estimators MTL95, MTL90 and MTL85.
In the middle row $\lambda$ is 1, and we see that
MTL95, which fits $95\%$ of the data, performs well
when there are up to $5\%$ of outliers and
performs poorly when there are over $5\%$ of 
outliers.
Analogously MTL90 performs well as long as there 
are at most $10\%$ of outliers, and so on.
This is the intended behavior.
But note that for $\lambda \neq 1$ these estimators 
also have a substantial bias when the fraction of
outliers is {\it below} what they aim for, as can 
be seen in the top and bottom panels of 
Figure \ref{fig:allepsplotsYJ}.
For instance MTL85 is biased when $\eps$ is under
$15\%$, even for $\eps = 0\%$ when there are no 
outliers at all. 
So overall the MTL estimators only performed well 
when the percentage of trimming was equal to 
1 minus the percentage of outliers in the data. 
Since the true percentage of outliers is almost 
never known in advance, it is not recommended to 
use the MTL method for variable transformation.

\begin{figure}[!ht]
\begin{centering}
\includegraphics[width = 0.82\textwidth]
  {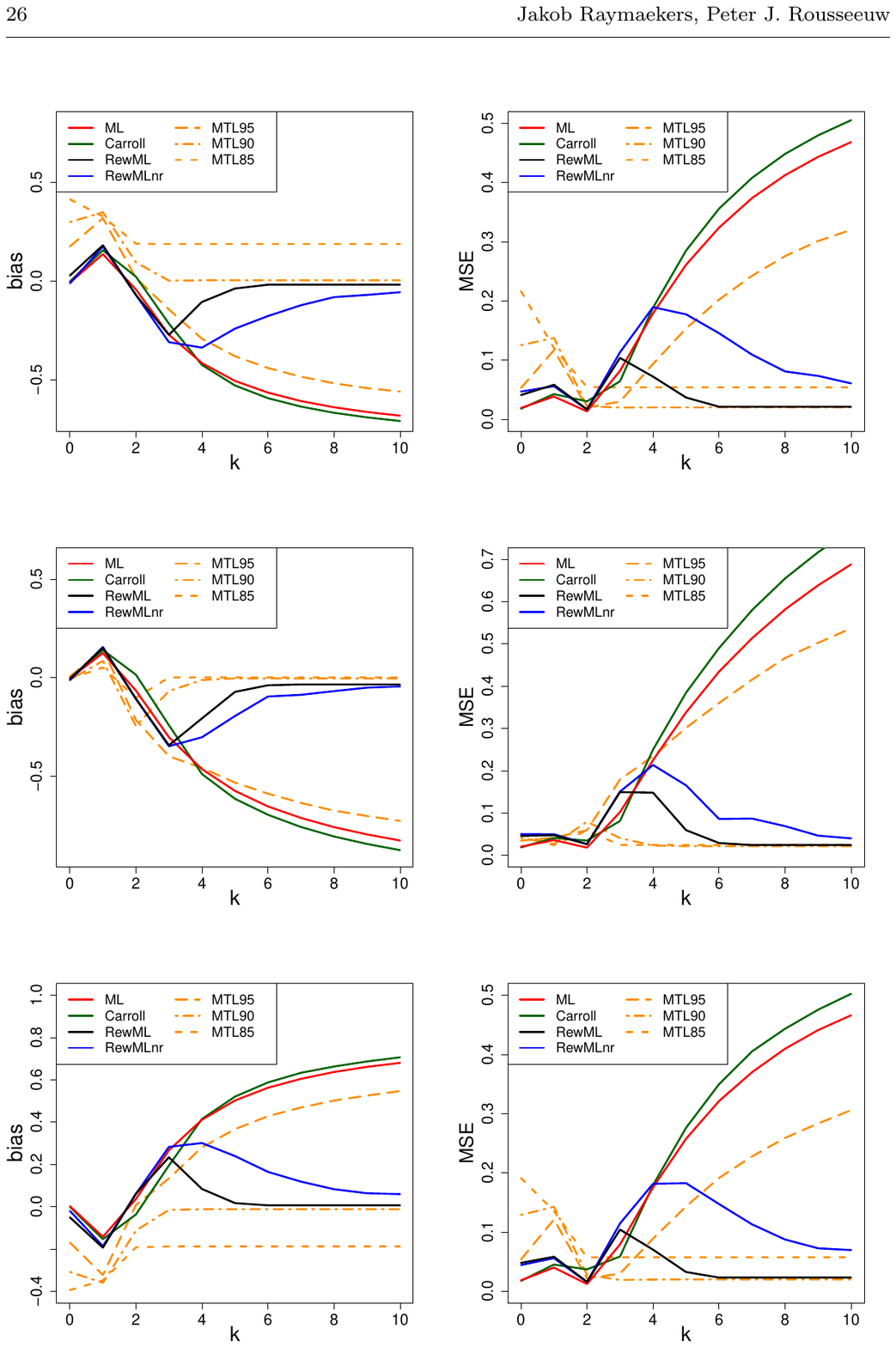}\\
\end{centering}
\caption{Bias (left) and MSE (right) of the
  estimated $\lambdahat$ of the
  Yeo-Johnson transformation as a function of $k$ 
	which determines how far the outliers are.
	Here the percentage of outliers is fixed at $10\%$\,. 
	The true parameter $\lambda$ used to generate
	the data is 0.5 in the top row, 1.0 in the
	middle row, and 1.5 in the bottom row.}
\label{fig:allkplotsYJ}
\end{figure}

Let us now investigate what happens if we 
keep the percentage of outliers fixed, say at 
$\eps = 10\%$\,, but vary the position of the 
contamination by letting $k = 0,1,\ldots,10$. 
Figure \ref{fig:allkplotsYJ} shows the resulting bias 
and MSE, with again $\lambda=0.5$ in the top row, 
$\lambda=1$ in the middle row,
and $\lambda=1.5$ in the bottom row.
For $k=0$ and $k=1$ the ML, Carroll, RewML and RewMLnr
methods give similar results, since the contamination 
is close to the center so it cannot be considered 
outlying.
But as $k$ increases the classical ML and the Carroll 
estimator become heavily affected by the outliers. 
On the other hand RewML and RewMLnr perform much 
better, and again RewML outperforms RewMLnr.
Note that the bias of RewML moves toward zero
when $k$ is large enough. We already noted this
redescending behavior in its sensitivity curve
in Figure \ref{fig:SC}.

The behavior of the maximum trimmed likelihood
methods depends on the value of $\lambda$
used to generate the data.
First focus on the middle row of Figure 
\ref{fig:allkplotsYJ} where $\lambda = 1$ so 
the clean data is generated from the standard normal 
distribution. 
In that situation both MTL90 and MTL85 behave well,
whereas MTL95 can only withstand $5\%$ of outliers
and not the $10\%$ generated here.
One might expect the MTL method to work well
as long as its $h$ excludes at least the number of
outliers in the data. 
But in fact MTL85 does not behave so well when
$\lambda$ differs from 1, as seen in the top and
bottom panels of Figure \ref{fig:allkplotsYJ},
where the bias remains substantial even though
the method expects 15\% of outliers and there are
only 10\% of them.
As in Figure \ref{fig:allepsplotsYJ} this suggests
that one needs to know the actual percentage of
outliers in the data in order to select the 
appropriate $h$ for the MTL method, but that 
percentage is typically unknown.

%%%%%%%%%%%%%%%%%%%%%%%%%%%%%%%%%%%%%%%%%%%%%%%%%%%
\subsection{Results for the Box-Cox transformation}
\label{sec:ResultsBC}

\begin{figure}[!ht]
\vskip0.3cm
\begin{centering}
\includegraphics[width = 0.83\textwidth]
  {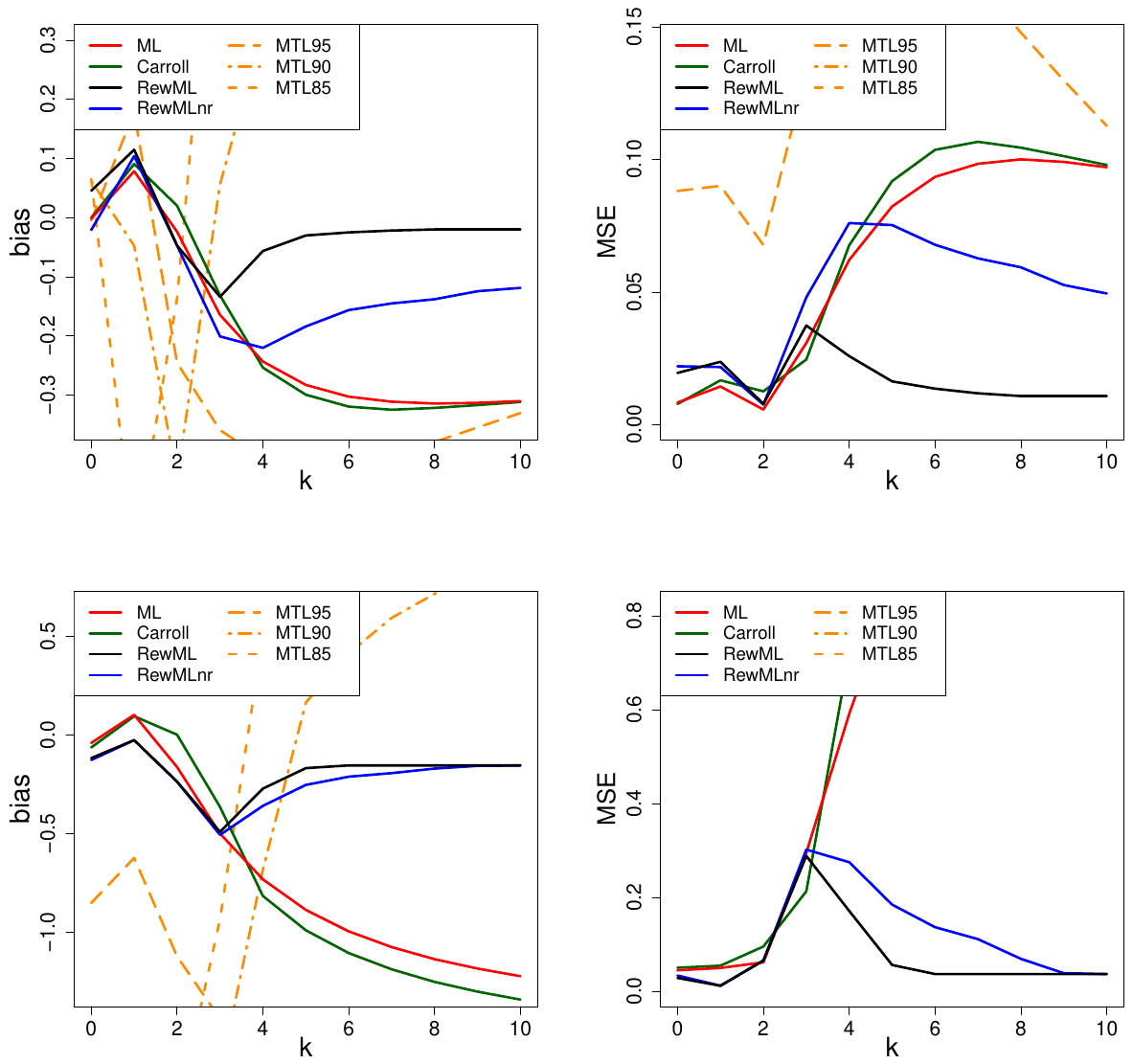}\\
\end{centering}
\caption{Bias (left) and MSE (right) of the
  estimated $\lambdahat$ of the 
  Box-Cox transformation as a function of $k$ which
	determines how far the outliers are.
	Here the percentage of outliers is fixed at $10\%$\,.
	The true $\lambda$ used to generate	the
	data is 0 in the top row and 1 in the bottom row.}
\label{fig:allkplotsBC}
\end{figure}

When simulating data to apply the Box-Cox
transformation to, the most natural choice of 
$\lambda$ is zero since this is the only value
for which the range of BC is the entire real line.
Therefore we can carry out the inverse BC 
transformation on any data set generated from a 
normal distribution, so the clean data follows a 
log-normal distribution.
The top panel of Figure \ref{fig:allkplotsBC}
shows the bias and MSE for $10\%$ of outliers
with $k=1,\ldots,10$. 
We see that the classical ML and the estimator of 
Carroll are sensitive to outliers when $k$ grows.
Our reweighted method RewML performs much better. 
The RewMLnr method only differs from RewML in that 
it uses the non-rectified BC transform in the first
step, and does not do as well since its bias goes
back to zero at a slower rate. 

The MTL estimators perform poorly here. 
The MTL95 version trims only $5\%$ of the data points
so it cannot discard the 10\% of outliers, leading
to a behavior resembling that of the classical ML.
But also MTL90 and MTL85, which trim enough data
points, obtain a large bias which goes in the 
opposite direction, accompanied by a high MSE.
The MSE of MTL90 and MTL85 lie entirely above the 
plot area.

Finally, we consider a scenario with 
$\lambda = 1$. In that case the range of the
Box-Cox transformation given by \eqref{eq:rangeBC}
is only $(-1,+\infty)$ so the transformed data
cannot be normally distributed (which is already
an issue for the justification of the classical 
maximum likelihood method).
But the transformed data can have a truncated
normal distribution.
In this special setting we generated data from the 
normal distribution with mean 1 and standard 
deviation $1/3$, and then truncated it to 
$[0.01, 1.99]$ (keeping $n=100$ points), so
the clean data are strictly positive and have
a symmetric distribution around 1.
In the bottom panel of Figure \ref{fig:allkplotsBC}
we see that the ML and Carroll estimators are not 
robust in this simulation setting. 
The trimmed likelihood estimators also fail to 
deliver reasonable results, with curves that
often fall outside the plot area. 
On the other hand RewML still performs well, and 
again does better than RewMLnr.\newline
\indent The simulation results for a fixed 
outlier position at $k=6$ or $k=10$ with 
contamination levels $\eps$ from 0\% to 15\% can 
be found in section B %\ref{secA:incc} 
of the supplementary material, and are qualitatively 
similar to those for the YJ transform.

%%%%%%%%%%%%%%%%%%%%%%%%%%%%%%%%%%%%%%%%%%%%%%%%%%%%
\section{Empirical examples}
\label{sec:real}

%%%%%%%%%%%%%%%%%%%%%
\subsection{Car data}

\begin{figure}[!ht]
\center
\includegraphics[width = 0.49\textwidth]
  {MPG_BC_ML_QQ_cr.pdf}
\includegraphics[width = 0.49\textwidth]
  {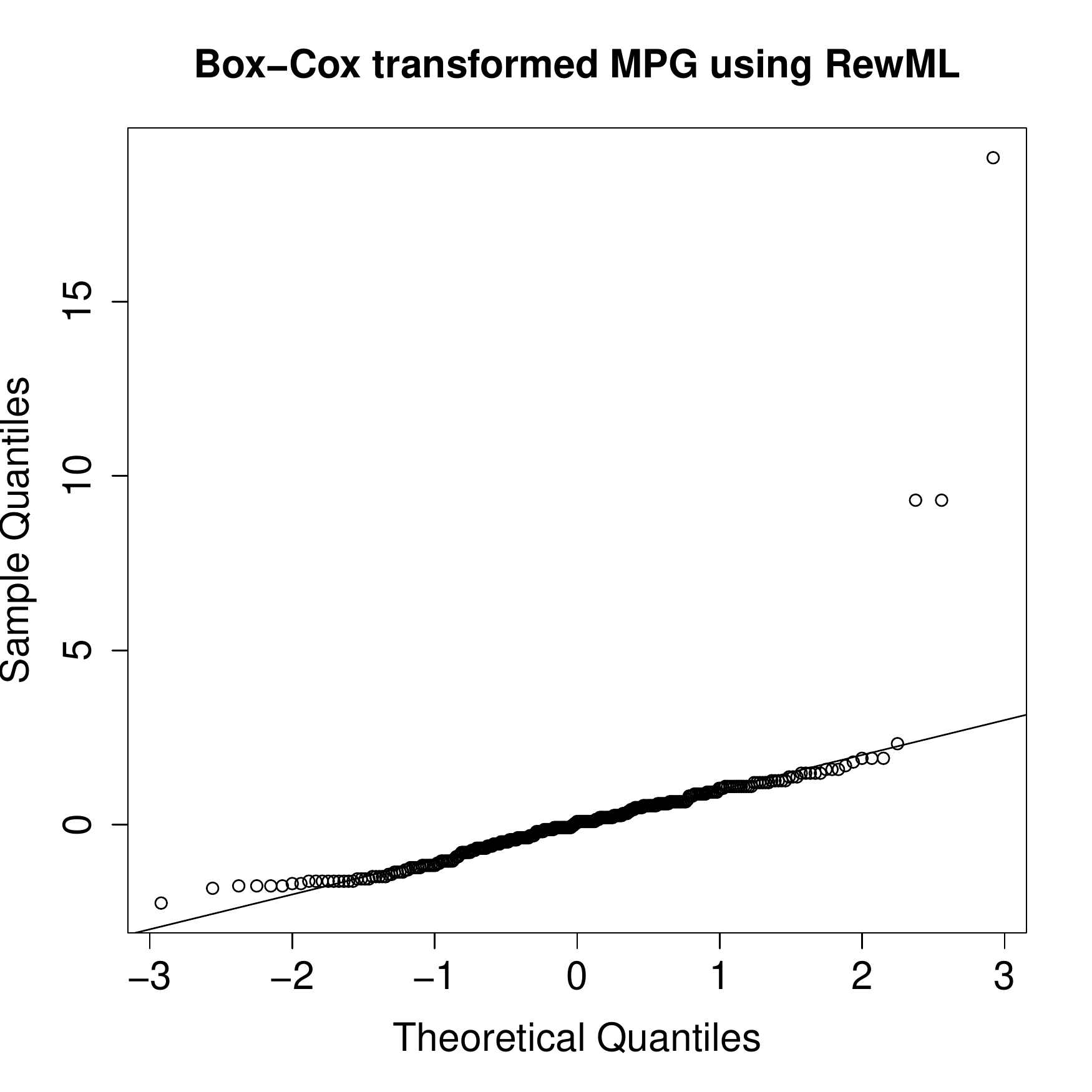}	
\caption{Normal QQ-plot of the Box-Cox transformed 
variable MPG using the ML estimate of $\lambda$ 
(left) and using the RewML estimate (right). 
The ML is strongly affected by the 3 outliers
at the top, thereby transforming the central data 
away from normality. 
The RewML method achieves central normality and 
makes the outliers stand out more.}
\label{Fig:MPGQQ2}
\end{figure}

Let us revisit the positive variable \texttt{MPG} 
from the TopGear data shown in the left panel of
Figure \ref{Fig:MPGQQ}.
The majority of these data are already roughly 
normal, and three far outliers at the top 
deviate from this pattern.
Before applying a Box-Cox transformation 
we first scale the variable so its median becomes 1.
This makes the result invariant to the unit of
measurement, whether it is miles per gallon or,
say, kilometers per liter.
The maximum likelihood estimator for Box-Cox 
tries to bring 
in the outliers and yields $\hat{\lambda} = -0.11$, 
which is close to $\lambda=0$ corresponding to a 
logarithmic transformation.
The resulting transformed data in the left panel 
of Figure \ref{Fig:MPGQQ2} are quite skewed in the 
central part, so not normal at all, which defeats 
the purpose of the transformation.
This result is in sharp contrast with our reweighted 
maximum likelihood (RewML) method which estimates the 
transformation parameter as $\hat{\lambda} = 0.84$. 
The resulting transformed data in the right
panel does achieve central normality.

\begin{figure}[!ht]
\center
\includegraphics[width = 0.49\textwidth]
  {Weight_BC_ML_QQ_cr.pdf}
\includegraphics[width = 0.49\textwidth]
  {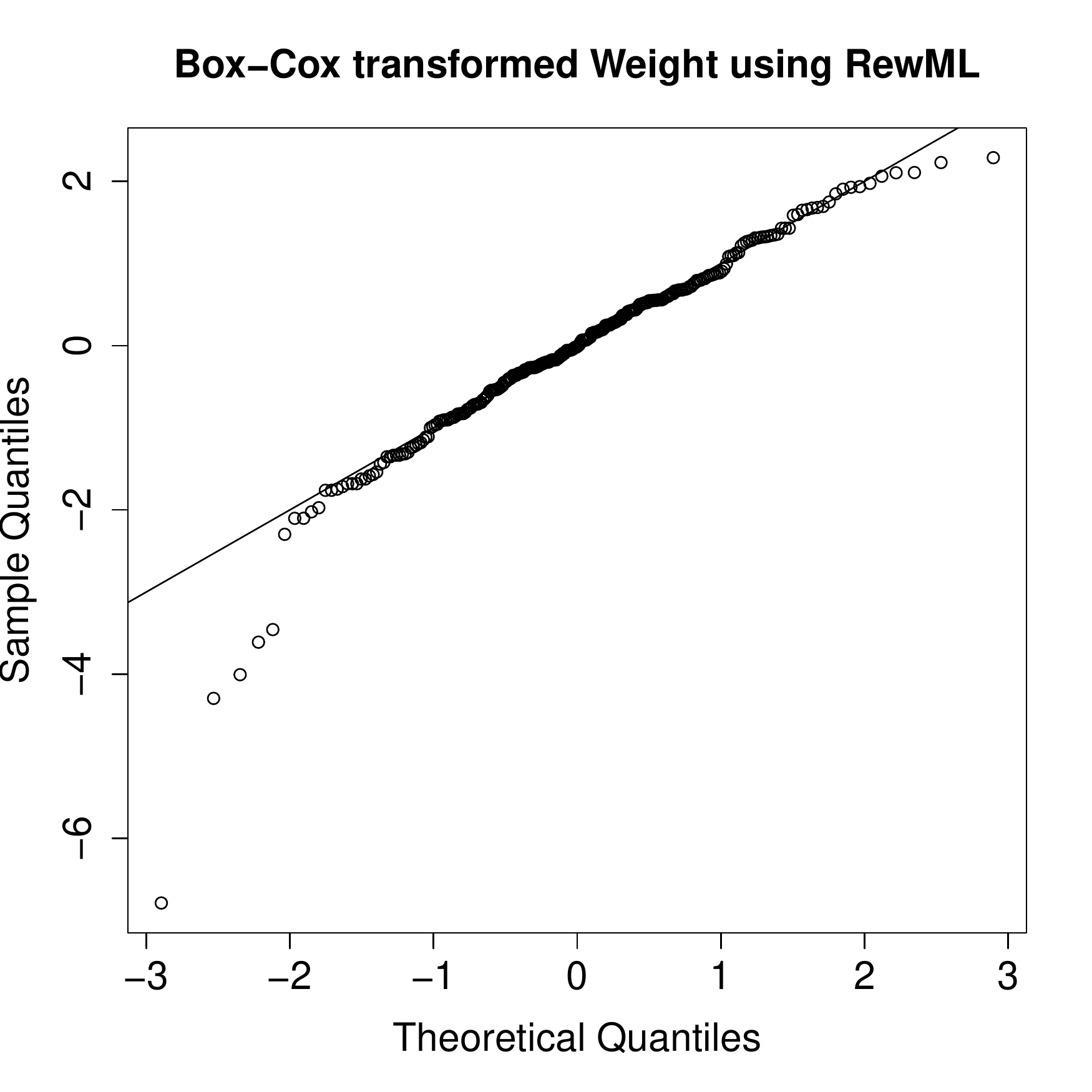}	
\caption{Normal QQ-plot of the Box-Cox transformed 
variable \texttt{Weight} using the ML estimate of 
$\lambda$ (left) and using the RewML estimate 
(right). The ML masks the five outliers at the
bottom, whereas RewML accentuates them and 
achieves central normality.}
\label{Fig:WeightQQ2}
\end{figure}

The variable \texttt{Weight} in the left
panel of Figure \ref{Fig:WeightQQ} is not very normal
in its center and has some outliers at the bottom.
The classical ML estimate is $\hat{\lambda} = 0.83$, 
close to $\lambda=1$ which would not transform
the data at all, as we can see in the resulting left 
panel of Figure \ref{Fig:WeightQQ2}. 
In contrast, our RewML estimator obtains 
$\hat{\lambda} = 0.09$ which substantially transforms 
the data, yielding the right panel of Figure 
\ref{Fig:WeightQQ2}.
There the central part of the data is very close to 
normal, and the outliers at the bottom now stand out 
more, as they should.

%%%%%%%%%%%%%%%%%%%%%%%
\subsection{Glass data}

For a multivariate example we turn to
the \texttt{glass} data 
\cite{Lemberge2000,DDC2018}
from chemometrics, which has become something of 
a benchmark.
The data consists of $n = 180$ archeological 
glass samples, which were analyzed by 
spectroscopy. Our variables are the intensities 
measured at 500 wavelengths.
Many of these variables do not look normally 
distributed.

\begin{figure}[!ht]
\center
\includegraphics[width = 0.8\textwidth]
  {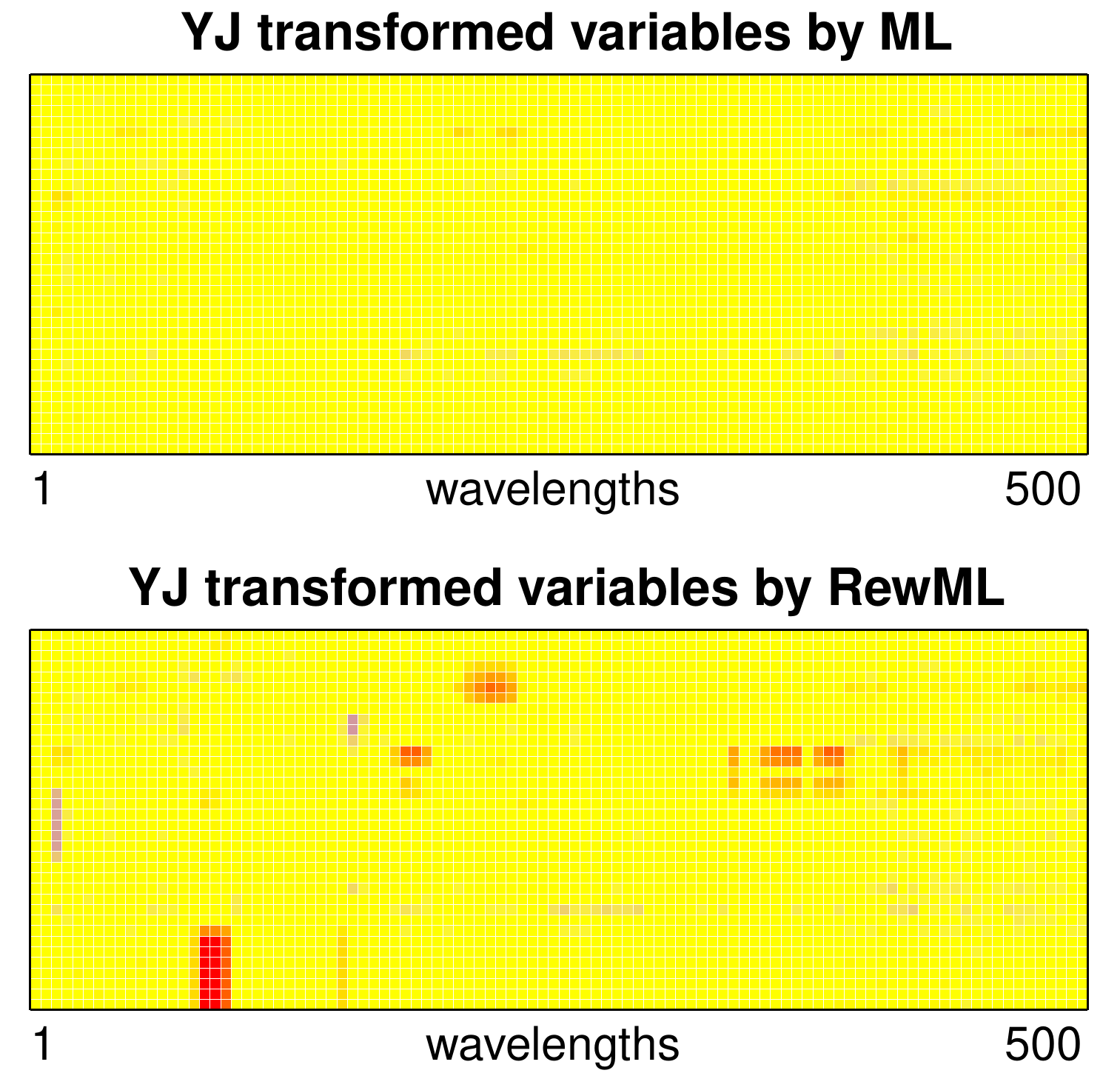}
\caption{Heatmap of the \texttt{glass} data after
transforming each variable (column) by a Yeo-Johnson 
transform with parameter $\lambda$ estimated by
(top) the maximum likelihood method, and (bottom)
the reweighted maximum likelihood method RewML. 
Yellow cells correspond to values in the central
$99\%$ range of the normal distribution.
Red cells indicate unusually high values, 
and blue cells have unusually low values.}
\label{fig:Glass}
\end{figure}

We first applied a Yeo-Johnson transformation
to each variable with $\lambdahat$ obtained
from the nonrobust ML method of \eqref{eq:ML}.
For each variable we then standardized 
the transformed data $\YJlhat(x_i)$ to
$(\YJlhat(x_i)-
 \muhat_{\mbox{\tiny{ML}},\lambdahat})/
 \sigmahat_{\mbox{\tiny{ML}},\lambdahat}$
where $\muhat_{\mbox{\tiny{ML}},\lambdahat}$
and $\sigmahat_{\mbox{\tiny{ML}},\lambdahat}$
are given by \eqref{eq:musigmaML}.
This yields a standardized transformed data
set with again 180 rows and 500 columns.
In order to detect outliers in this matrix
we compare each value to the interval
$[-2.57,2.57]$ which has a probability of
exactly $99\%$ for standard normal data.
The top panel of Figure \ref{fig:Glass} is
a heatmap of the standardized transformed data 
matrix where each value within $[-2.57,2.57]$
is shown as yellow, values above $2.57$ are 
red, and values below $-2.57$ are blue.
This heatmap is predominantly yellow because
the ML method tends to transform the data in
a way that masks outliers, so not much
structure is visible.

Next, we transformed each variable by Yeo-Johnson 
with $\lambdahat$ obtained by the robust RewML 
method. The transformed variables were 
standardized accordingly to $(\YJlhat(x_i)-
 \muhat_{\mbox{\tiny{W}},\lambdahat})/
 \sigmahat_{\mbox{\tiny{W}},\lambdahat}$
where $\muhat_{\mbox{\tiny{W}},\lambdahat}$
and $\sigmahat_{\mbox{\tiny{W}},\lambdahat}$
are given by \eqref{eq:musigmaW} using the
final weights in \eqref{eq:WMLYJ}.
The resulting heatmap is in the bottom panel
of Figure \ref{fig:Glass}.
Here we see much more structure, with red
regions corresponding to glass samples with
unusually high spectral intensities at certain 
wavelengths.
This is because the RewML method aims to make
the central part of each variable as normal
as possible, which allows outliers to deviate 
from that central region. 
The resulting heatmap has a subject-matter
interpretation since wavelengths correspond 
to chemical elements.
It indicates that some of the glass samples 
(with row numbers between 22 and 30) have a 
higher concentration of phosphor, whereas rows 
57–63 and 74–76 had an unusually high amount 
of calcium. The red zones in the bottom part 
of the heatmap were caused by the fact that the 
measuring instrument was cleaned before 
recording the last 38 spectra.

%%%%%%%%%%%%%%%%%%%%%%%
\subsection{DPOSS data}

As a final example we consider data from the
Digitized Palomar Sky Survey (DPOSS) described by
\cite{djorgovski1998}. We work with the dataset
of 20000 stars available as \texttt{dposs} in the 
R package \texttt{cellWise} \cite{Raymaekers2020}.
The data are measurements in three color bands,
but for the purpose of illustration we restrict
attention to the color band with the fewest
missing values. Selecting the completely observed
rows then yields a dataset of 11478 observations
with 7 variables. Variables MAper, MTot and
MCore measure light intensity, and variables
Area, IR2, csf and Ellip are measurements of 
the size and shape of the images.

\begin{figure}[!ht]
\center
\includegraphics[width = 0.75\textwidth]
   {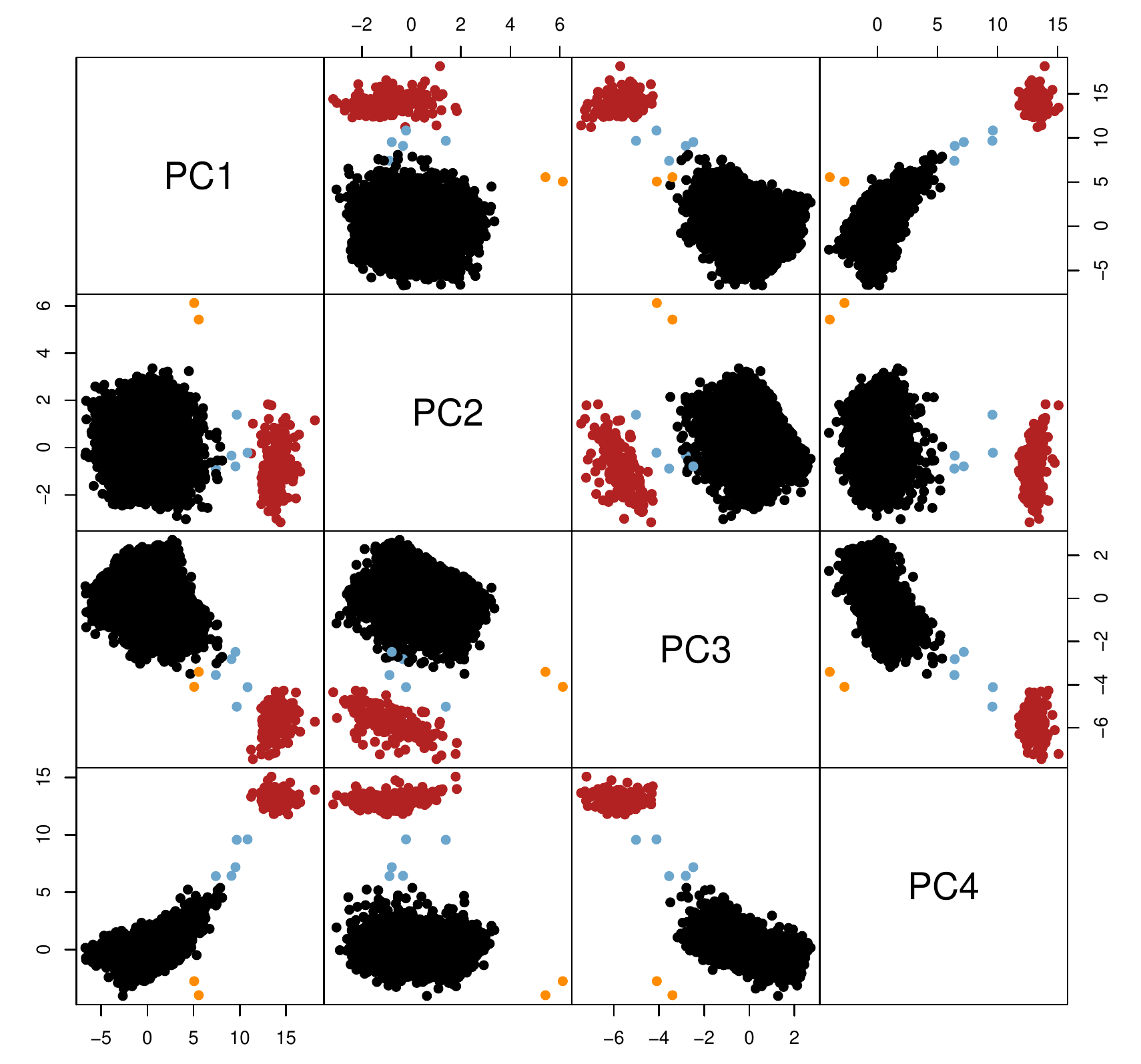} 
\caption{DPOSS data: pairs plot of the robust PCA 
  scores after transforming the data with the 
	robustly fitted YJ-transformation.}
\label{fig:DPOSS_scores}
\end{figure}

In order to analyze the data we first apply 
the YJ-transformation to each variable, with the
$\lambdahat$ estimates obtained from RewML.
We then perform cellwise robust PCA 
\cite{hubert2019macropca}. 
We retained
$k=4$ components, explaining 97\% of the variance. 
Figure \ref{fig:DPOSS_scores} is the pairs plot of 
the robust scores.  
We clearly see a large cluster of regular points
with some outliers around it, and a smaller cluster
of rather extreme outliers in red. The red points
correspond to the stars with a high value of
MAper. The blue points are somewhat in between
the main cluster and the smaller cluster of extreme
outliers. The two orange points are celestial
objects with an extreme value of Ellip.

\begin{figure}[!ht]
\center
\includegraphics[width = 0.495\textwidth]
   {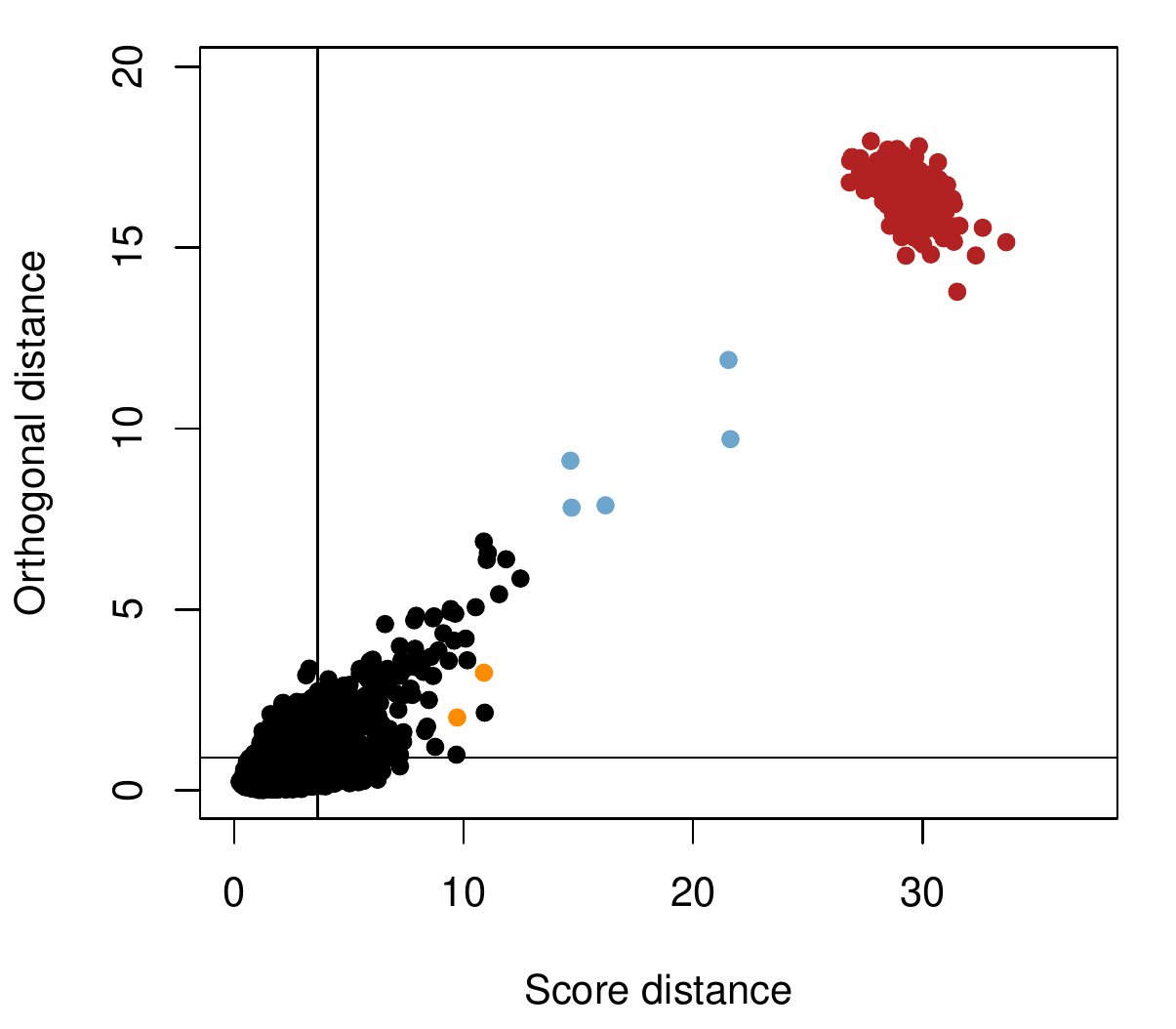}
\includegraphics[width = 0.495\textwidth]
   {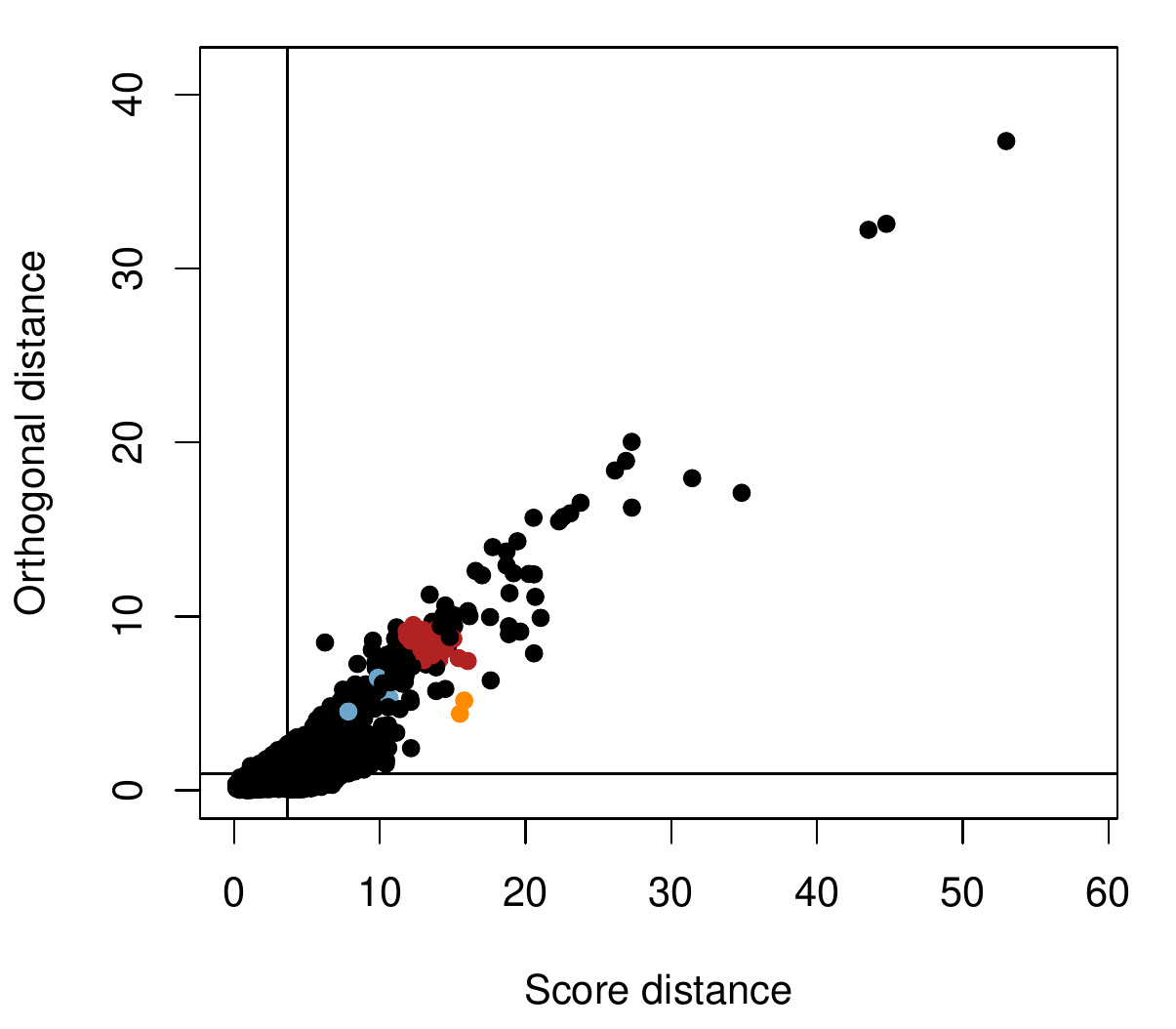}	
\caption{DPOSS data: outlier map of robust PCA 
  applied to the YJ-transformed variables (left), 
	and applied to the raw variables (right).
	The colors are those of Figure 
	\ref{fig:DPOSS_scores}.}
\label{fig:DPOSS_outlmaps}
\end{figure}

The left panel of Figure 
\ref{fig:DPOSS_outlmaps} contains the 
{\it outlier map} \cite{ROBPCA2005}
of this robust PCA. 
Such an outlier map consists of two ingredients.
The horizontal axis contains the 
{\it score distance} of each object $\bx_i$.
This is the robust Mahalanobis distance of the
\mbox{orthogonal} projection of $\bx_i$ on the subspace 
spanned by the $k$ principal components, and can 
be computed as
\begin{equation} \label{eq:SD}
  \SD_i = \sqrt{\sum_{j=1}^{k} 
	         \frac{t_{ij}^2}{\lambda_j} }  
\end{equation}
where $t_{ij}$ are the PCA scores and $\lambda_j$
is the $j$-th largest eigenvalue.
The vertical axis shows the {\it orthogonal 
distance} $\OD_i$ of each point $\bx_i$\,, which is 
the distance between $\bx_i$ and its projection 
on the $k$-dimensional subspace.
We see that the red points form a large cluster 
with extreme $\SD_i$ as well as $\OD_i$.
The blue points are intermediate, and the two
orange points are still unusual but less extreme.

Interestingly, when applying robust PCA to the 
untransformed data we obtain the right panel
of Figure \ref{fig:DPOSS_outlmaps} which shows
a very different picture in which the red, blue
and orange points are not clearly distinguished 
from the remainder. 
This outlier map appears to indicate a number of
black points as most outlying, but in fact they
should not be considered outlying since they 
are characterized by high values of Area and 
IR2 which are very right-skewed. 
When transforming these variables to central 
normality, as in the left panel, the values that
appeared to be outlying become part of the regular
tails of a symmetric distribution.

%%%%%%%%%%%%%%%%%%%%%%%%%%%%%%%%%%%%%%%%%%%%%%%%%
\section{Discussion}
\label{sec:disc}
In this discussion we address and clarify a few 
aspects of (robust) variable transformation. 

{\bf Outliers in skewed data.} 
The question of what is an outlier in a general
skewed distribution is a rather difficult one. 
One could flag the values below a lower cutoff or
above an upper cutoff, for instance given by
quantiles of a distribution. 
But it is hard to robustly determine cutoffs
for a general skewed dataset or distribution. 
Our viewpoint is this. 
If the data can be robustly transformed by BC or
YJ towards a distribution that looks normal except 
in the tails, then it easy to obtain cutoffs for
the transformed data, e.g. by taking their median
plus or minus two median absolute deviations.
And then, since the BC and YJ transforms are
monotone, these cutoffs can be transformed back
to the original skewed data or distribution.
In this way we do not flag too many points. 
We have illustrated this for lognormal data in 
section C % \ref{secA:outldet} 
of the supplementary material.

{\bf How many outliers can we deal with?}
In addition to sensitivity curves, which
characterize the effect of a single outlier, it is
interesting to find out how many adversely placed
outliers it takes to make the estimators fail badly. 
For RewML this turns out to be around 15\% of the 
sample size, which is quite low compared to robust 
estimators of location or scale. 
But this is unavoidable, as asymmetry is tied to 
the tails of the distribution. 
Empirically, skewness does not manifest itself 
much in the central part of the distribution, 
say 20\% of mass above and below the median. 
On each side we have 30\% of mass left, and if the
adversary is allowed to replace half of either portion
(that is, 15\% of the total mass) and move it 
{\it anywhere}, they can modify the skewness 
a great deal.

{\bf Prestandardization.}
In practice, one would typically apply some form of 
prestandardization to the data before trying
to fit a transformation. 
For the YJ transformation, we can simply
prestandardize the dataset $X = \{x_1,\ldots,x_n\}$ by
\begin{equation} \label{eq:prestandYJ}
  \tilde{x}_i = \frac{x_i - \mbox{median}(X)}
	              {\mbox{mad}(X)}
\end{equation}
where $\mbox{mad}$ is the median absolute
deviation from the median.
This is what we did in the glass and DPOSS examples.
An advantage of this prestandardization is that
the relevant $\lambda$ will typically be in the 
same range (say, from -4 to 6) for all variables
in the data set, which is convenient for the
computation of $\lambdahat$.
Before the BC transformation we cannot prestandardize 
by \eqref{eq:prestandYJ} since this would generate 
negative values. One option is to
divide the original $x_i$ by their median so they 
remain positive and have median 1.
But if the resulting data is tightly concentrated
around 1, we may still require a huge positive or 
negative $\lambda$ to properly transform them by BC.
Alternatively, we propose to prestandardize by
\begin{equation} \label{eq:prestandBC}
  \tilde{x}_i = \exp\Big(\frac{\log(x_i) - 
	 \mbox{median}(\log X)} {\mbox{mad}(\log X)}\Big)
\end{equation}
which performs a standardization on the log scale
and then transforms back to the positive scale.
This again has the advantage that the range of 
$\lambda$ can be kept fixed when computing
$\lambdahat$\,, making it as easy as applying the 
YJ transform after \eqref{eq:prestandYJ}.
A disadvantage is that the transformation
parameter becomes harder to interpret, since e.g.
a value of $\lambda = 1$ no longer corresponds to
a linear transformation, but $\lambda = 0$ still 
corresponds to a logarithmic transform.

{\bf Tuning constants.} 
The proposed method has some tuning parameters, 
namely the constant $c$ in \eqref{eq:rhobw}, the 
cutoff $\Phi^{-1}(0.995)$ in the reweighting step, 
and the rectification points.
The tuning of the constant $c$ is addressed in
section A % \ref{secA:tuningc} 
of the supplementary
material. The second parameter is the constant 
$\Phi^{-1}(0.995) \approx 2.57$ in the weights 
\eqref{eq:weights}.
This weight function is commonly used in existing
reweighting techniques in robust statistics. 
If the transformed data is normally distributed we 
flag approximately 1\% of values with this cutoff,
since $P(|Z| > \Phi^{-1}(0.995)) = 0.01$ if
$Z \sim \mathcal{N}(0,1)$. Choosing a higher cutoff
results in a higher efficiency, but at the cost of
lower robustness. Generally, the estimates do not
depend too much on the choice of this cutoff as long
as it is in a reasonably high range, say from
$\Phi^{-1}(0.975)$ to $\Phi^{-1}(0.995)$.
A final choice in the proposal are the
``rectification points'' $C_\ell$ and $C_u$ for which
we take the first and third quartiles of the data.
Note that the constraints $C_\ell < 0 < C_u$ for YJ
and $C_\ell < 1 < C_u$ for BC in 
\eqref{eq:BCright}--\eqref{eq:YJleft} are satisfied
automatically when prestandardizing the data as
described in the previous paragraph. It is worth
noting that these data-dependent choices of $C_\ell$
and $C_u$ do not create
abrupt changes in the transformed data when moving
through the range of possible values for $\lambda$
because the rectified transformations are 
continuous in $\lambda$ by
construction. For instance, passing from 
$\lambda < 1$ to $\lambda > 1$ does not cause a 
jump in the transformed data because $\lambda=1$
corresponds to a linear transformation that is 
inherently rectified on both sides.

{\bf Models with transformed variables.}
The ease or difficulty of interpreting a model in 
the transformed variables depends on the model under 
consideration. For nonparametric models it makes
little difference.
For parametric models the transformations can make 
the model harder to interpret in some cases and easier
in others, for instance where there is a simple
linear relation in the transformed variables
instead of a model with higher-order terms in the
original variables. 
Also, the notion of leverage point in
linear regression is more easily interpretable with 
roughly normal regressors, as it is related to the 
(robust) Mahalanobis distance of a point in regressor 
space.

The effect of the sampling variability of
$\lambda$ on the inference in the resulting model
is also model dependent. We expect that it will
typically lead to a somewhat higher variability in 
the estimation. However, as the BY and YJ
transformations are continuous and differentiable 
in $\lambda$ and not very sensitive to small changes 
of $\lambda$, the increase in variability is likely
small. Moreover, if we apply BC or YJ 
transformations to predictor variables used in CART 
or Random Forest, the predictions and feature 
importance metrics stay exactly the same because 
the BC and YJ transformations are monotone.

%%%%%%%%%%%%%%%%%%%%%%%%%%%%%%%%%%%%%%%%%%%%%%%%%
\section{Conclusion}
\label{sec:conc}

In our view, a transformation to normality should
fit the central part of the data well, and not be
determined by any outliers that may be present.
This is why we aim for central normality, where
the transformed data is close to normal (gaussian) 
with the possible exception of some outliers that 
can remain further out.
Fitting such a transformation is not easy, because
a point that appears to be outlying in the original 
data may not be outlying in the transformed data, 
and vice versa.

To address this problem we introduced a combination 
of three ideas: a highly robust objective function
\eqref{eq:obj}, the rectified Box-Cox and 
Yeo-Johnson transforms in subsection 
\ref{sec:rectif} which we use in our initial 
estimator only, and a reweighted maximum 
likelihood procedure for transformations.
This combination turns out to be a powerful tool
for this difficult problem.

Preprocessing real data by this tool paves the
way for applying subsequent methods, such as anomaly 
detection and well-established model fitting and 
predictive techniques.

\vskip0.7cm

\noindent {\bf Software availability}

\noindent 
The proposed method is available as the
function \texttt{transfo()} in the 
\texttt{R} \cite{R2020} package 
\texttt{cellWise} \cite{Raymaekers2020} on 
CRAN, which also includes the vignette
\texttt{transfo\_examples} that reproduces all
the examples in this paper.
It makes uses of the \texttt{R}-packages 
\texttt{gridExtra} \cite{Auguie2017}, 
\texttt{reshape2} \cite{Wickham2007}, 
\texttt{ggplot2} \cite{Wickham2016} and 
\texttt{scales} \cite{Wickham2020}.\\ 

\noindent \textbf{Acknowledgement.} 
This work was supported by grant C16/15/068
of KU Leuven.

%\end{document} % for paper, not arXiv

\clearpage
%\pagenumbering{arabic}
   %% restarts page numbering from 1
	 %% Needed for paper, not for arXiv.
\appendix

\Large
\noindent \textbf{Supplementary Material} 
\normalsize

\section{Choice of the tuning constant c}
\label{secA:tuningc}
The function $\rhobw$ in \eqref{eq:rhobw} contains 
a tuning constant $c$ which we set at 0.5.
This tuning constant quantifies a trade-off between
variance and robustness. For smaller values of $c$
the central part of Tukey's $\rho$-function becomes
more narrow, making it less likely that clean
observations fall into this central part. For larger
values of $c$ the opposite happens. Ideally,
we would like the outliers to fall outside of the
central region, and as many of the inliers as possible
inside the central region. 
In section~\ref{sec:disc} it was argued that
robustly fitting a skewness-correcting transform
like BC or YJ can handle at most about 15\% of 
outliers.
We have tuned $c$ such that with 15\% of extreme 
contamination, approximately 90\% of the clean data
fall into the central region.

The computation goes as follows. Suppose we have a
data set $x_1, \ldots,x_n$ of size $n$ and assume 
w.l.o.g. that $x_1 \leq x_2 \leq \ldots \leq x_n$. 
Assume furthermore that a fraction $1-\eps$ can
be transformed to a gaussian distribution, but the 
remaining fraction $\eps$ consists of extreme 
outliers: $x_i \approx \infty$ for
$i=(1-\eps)n+1,\ldots,n$. 
(The situation where $x_i \approx -\infty$ for
$i=1,\ldots,\eps n$ is analogous.)
For the regular observations, the 
differences between the observed and 
expected quantiles in the objective function 
\eqref{eq:obj} are close to
$$ \Delta_i = \Phi^{-1}\left(\frac{i-1/3}
  {(1 - \eps)n+1/3}\right)- 
	\Phi^{-1}\left(\frac{i-1/3}{n+1/3}\right)$$	
for $i = 1,\ldots,(1-\eps)n$. We can verify that
$P(|\Delta| \leq 0.5) \approx 88\%$ when 
$n \geqslant 100$.

\section{Additional simulations}
\label{secA:incc}
In this section we show some additional simulation
results.
The plots complement the figures shown in the main
text by considering a different position of the
outliers ($k=6$) for an increasing contamination 
level $\eps$, and also the case of $k=10$ for
the BC transformation. The results shown here are
in line with the conclusions of the main text.

\subsection{the BC transformation}
\begin{figure}[!ht]
\begin{centering}
\includegraphics[width = 0.83\textwidth]
   {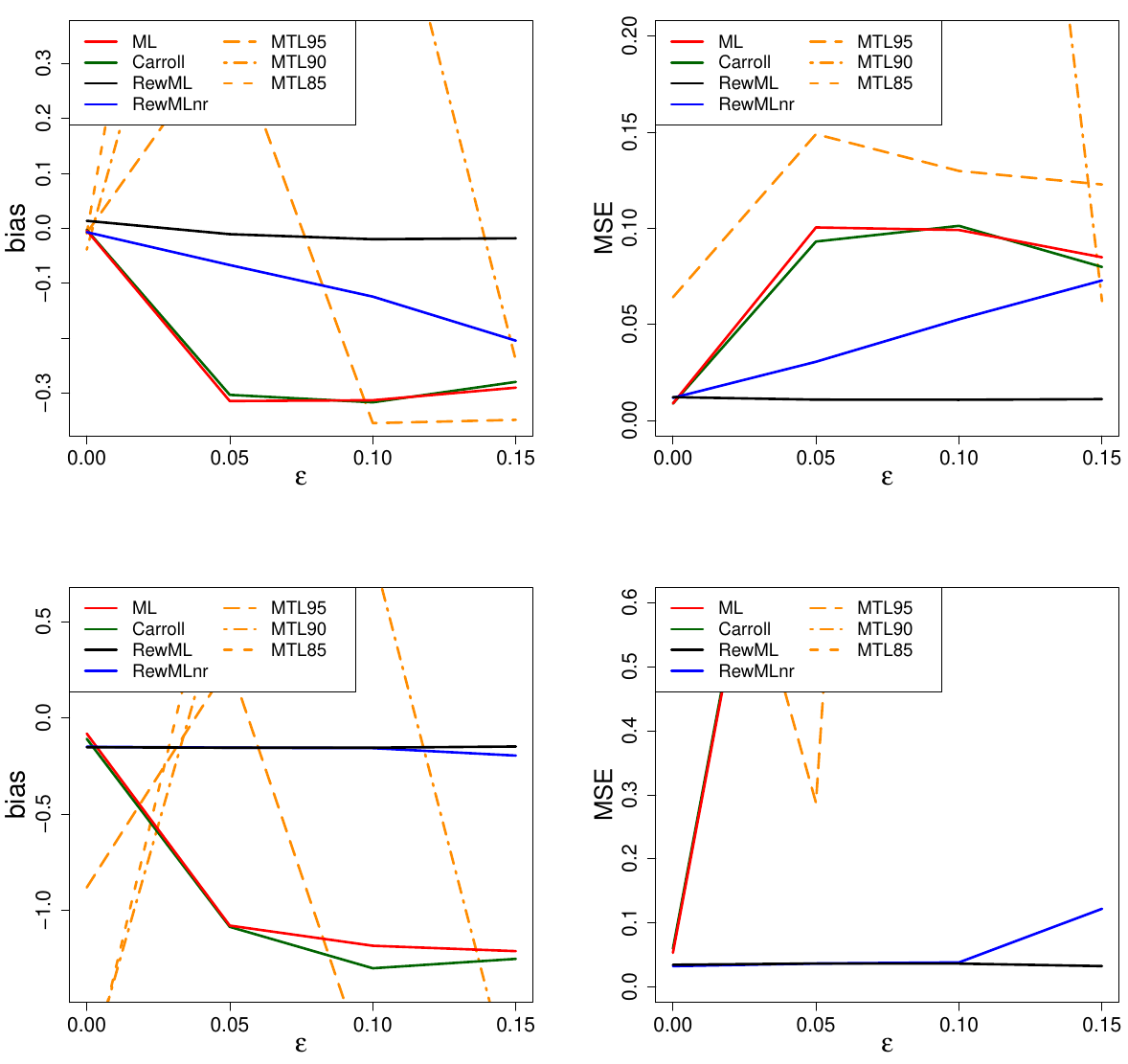}\\
\end{centering}	
\caption{Bias (left) and MSE (right) of the
  estimated $\lambdahat$ of the 
  Box-Cox transformation as a function of the 
	percentage $\eps$ of outliers, when the location 
	of the outliers is determined by setting $k=10$.
  The true parameter $\lambda$ used to generate
	the data is 0 in the top row 1 in the bottom row.}
\label{fig:allepsplotsBCk10}
\end{figure}

Figure \ref{fig:allepsplotsBCk10} shows the effect
of an increasing level of contamination on the
estimates of the BC-transformation parameter when
the position of the contamination is fixed at
$k = 10$. The proposed RewML did well,
as it has a very stable performance which is not 
affected much by the increasing level of 
contamination. The reweighted estimator starting 
from the initial estimator without rectification 
(RewMLnr) performs less well, especially when the 
percentage of contamination increases. 
Carroll's method and the classical maximum 
likelihood estimator are more heavily affected by 
the outliers, and perform poorly.
The trimmed likelihood estimators display a very
erratic behavior and are unreliable in this setup.

\begin{figure}[!ht]
\begin{centering}
\includegraphics[width = 0.83\textwidth]
   {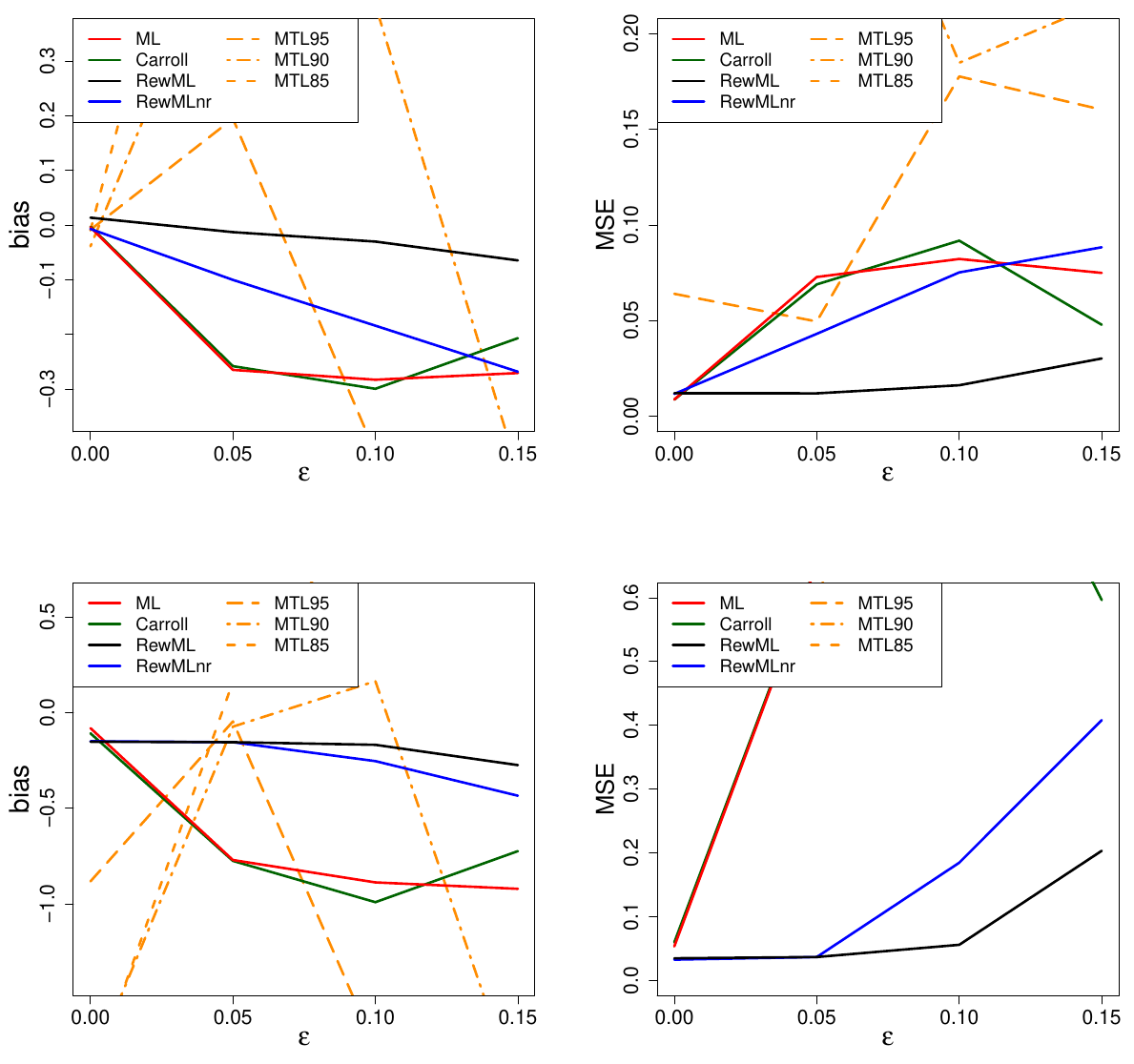}\\
\end{centering}	
\caption{Bias (left) and MSE (right) of the
  estimated $\lambdahat$ of the 
  Box-Cox transformation as a function of the 
	percentage $\eps$ of outliers, when the location 
	of the outliers is determined by setting $k=6$.
  The true parameter $\lambda$ used to generate
	the data is 0 in the top row, and 1 in the 
	bottom row.}
\label{fig:allepsplotsBCk6}
\end{figure}

Figure \ref{fig:allepsplotsBCk6} shows the effect
of an increasing level of contamination on the
estimates of the BC-transformation parameter when
the position of the contamination is fixed at $k=6$.
The general conclusions based from this figure are
the same as for the case $k = 10$ in the main text. 
We note that $k=6$ is slightly more difficult than
$k=10$ in Figure \ref{fig:allepsplotsBCk10}, as 
evidenced by the more pronounced effect of 
increasing the contamination level.

\FloatBarrier

\subsection{The YJ transformation}

\begin{figure}[!ht]
\begin{centering}
\includegraphics[width = 0.83\textwidth]
   {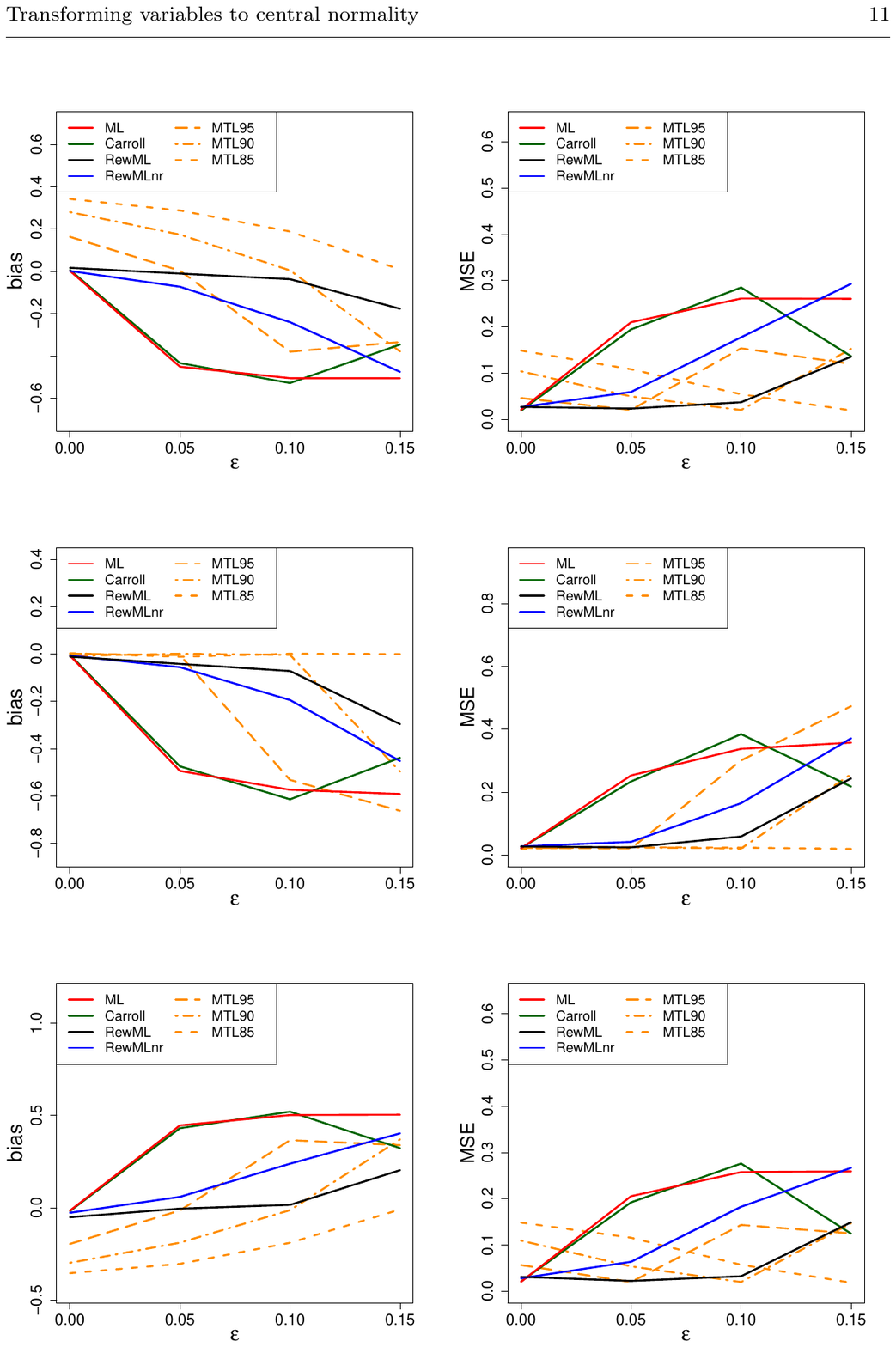}\\	
\end{centering}
\caption{Bias (left) and MSE (right) of the
  estimated $\lambdahat$ of the 
  Yeo-Johnson transformation as a function of the 
	percentage $\eps$ of outliers, when the location 
	of the outliers is determined by setting $k=6$.
  The true parameter $\lambda$ used to generate
	the data is 0.5 in the top row, 1.0 in the 
	middle row, and 1.5 in the bottom row.}
\label{fig:allepsplotsYJk6}
\end{figure}

Figure \ref{fig:allepsplotsYJk6} shows the effect
of an increasing level of contamination on the
estimates of the YJ transformation parameter when
the position of the contamination is fixed at $k=6$.
The performance of the estimators in this setting
is very similar to the case of $k = 10$, as can be
seen in a comparison with Figure 
\ref{fig:allepsplotsYJ} of the main text.
The trimmed estimators behave as they are supposed
to for $\lambda = 1$, but they become unreliable 
as soon as $\lambda$ deviates from 1. 
The ML estimator and Carroll's estimator are more 
heavily influenced by the outliers, but the 
magnitude of the effect is slightly smaller
than for $k=10$. The proposed RewML
estimator performs best, and we again see
that starting with an initial estimator without
rectification (RewMLnr) substantially reduces the
robustness of the reweighted estimator.

\clearpage
\section{False detections}
\label{secA:outldet}
\begin{figure}[!h]
\centering
\includegraphics[width = 0.6\textwidth]
   {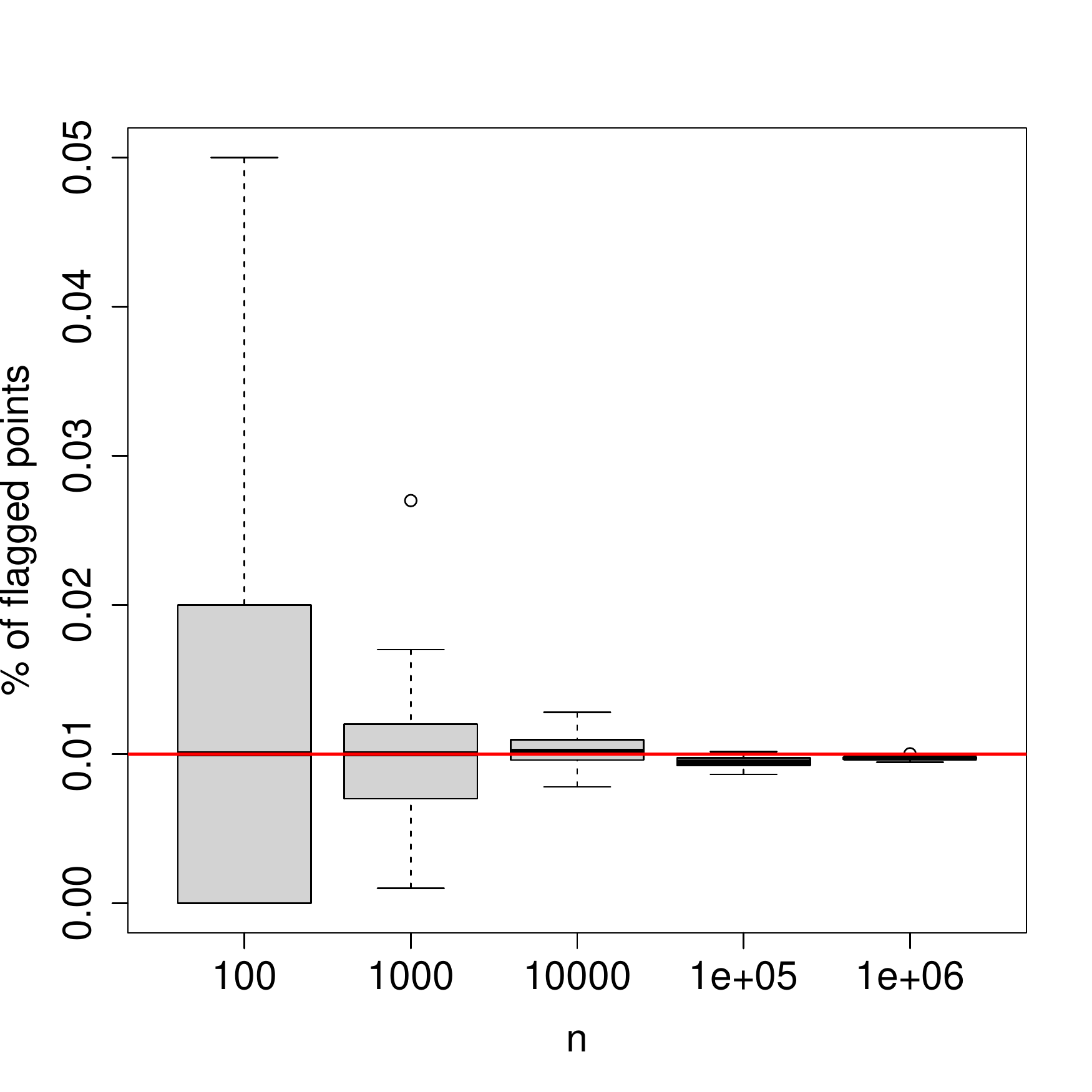}
\caption{Percentage of points flagged as outlying
in clean lognormal data, for various sample sizes.
As the sample size increases, the fraction of flagged
points approaches the expected frequency of 1\%.} 
\label{fig:outldet}
\end{figure}

We briefly investigate whether the proposed method is
well-calibrated when it comes to outlier detection.
More specifically, we track the number of
``false positives'', i.e. the number of observations
flagged as outliers when the data comes from a clean
distribution. The clean data was generated from the
lognormal distribution and the sample size was given 
by $n = 10^j$ for $j=2, \ldots, 6$. 
For each sample size 100 replications were run.
These uncontaminated data were Box-Cox transformed
with $\lambdahat$ obtained from the proposed RewML 
estimator.
Figure \ref{fig:outldet} shows boxplots of the 
percentage of points whose weight given by
\eqref{eq:weights} is zero. 
As the sample size $n$ increases the fraction of 
flagged points approaches 1\%, which is in line with 
the expected number of flagged points since the  
cutoffs on the transformed data in \eqref{eq:weights} 
are at the 0.005 and 0.995 quantiles.

%\clearpage
\section{Results with and without reweighting}
\label{secA:initial}
In this section we illustrate the importance of the
reweighting steps in RewML. Switching them off 
yields the initial estimator, which was left out
of the figures in the main text because
it fits a rectified transform, so it is 
effectively estimating a different transformation.
Therefore, a direct comparison between the initial 
rectified estimator and the other estimators
is questionable. 

\begin{figure}[!ht]
\centering
\includegraphics[width = 0.58\textwidth]
  {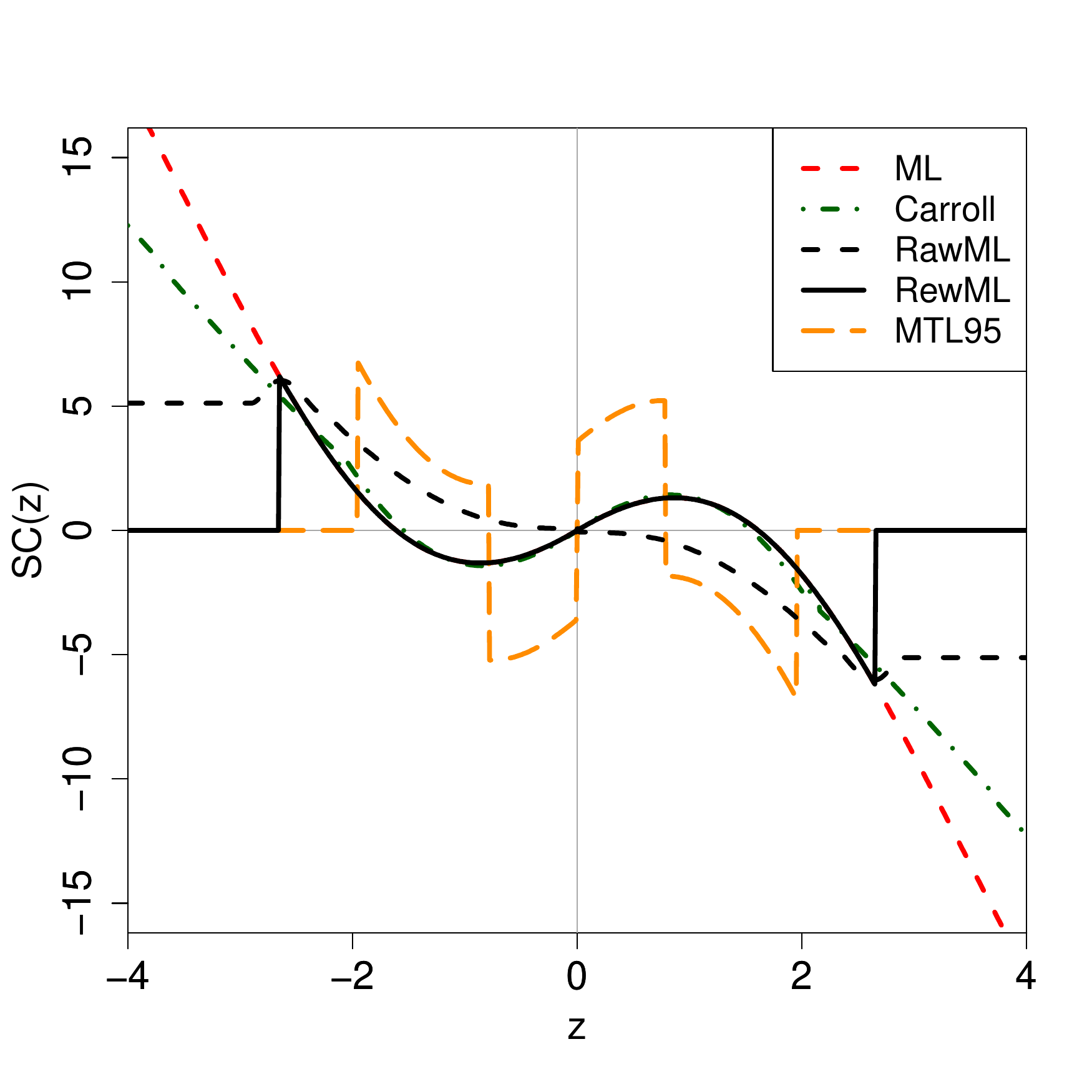}
\includegraphics[width = 0.58\textwidth]
  {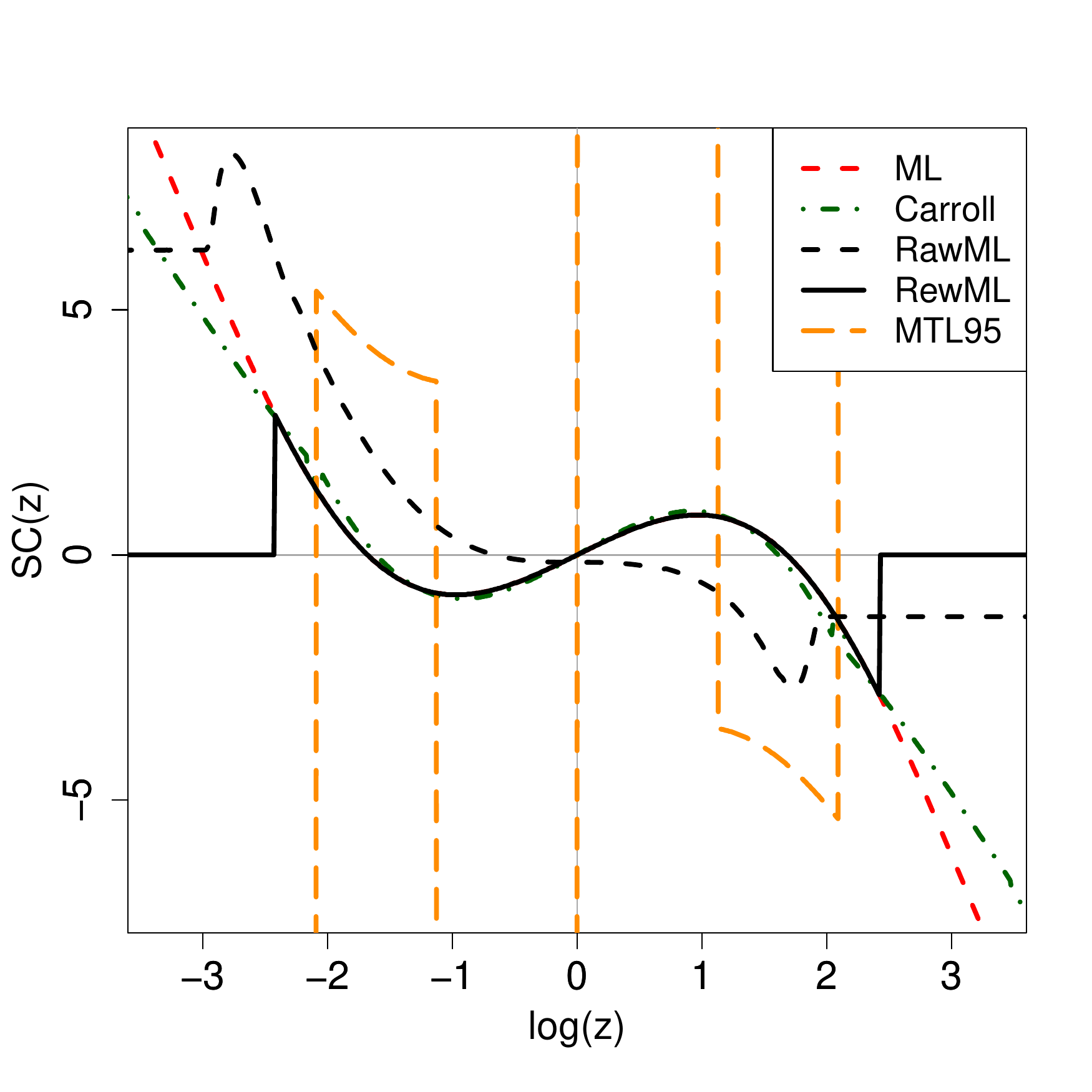}
\caption{Sensitivity curves of estimators of the 
  parameter $\lambda$ in the Yeo-Johnson 
	(top) and Box-Cox (bottom)
	transformations, with	sample size $n = 100$.}
\label{fig:SC_wraw}	
\end{figure}	

Figure \ref{fig:SC_wraw} shows the sensitivity 
curves including those of the initial
estimator, denoted as RawML. 
We see that RawML has a bounded sensitivity curve, 
but it does not redescend to zero for large values 
of $|z|$. The sensitivity curve of the
reweighted estimator behaves like the maximum
likelihood estimator in the center, but is zero for
large values of $|z|$. 
In the Box-Cox setting (lower panel of Figure 
\ref{fig:SC_wraw}) the sensitivity curve of RawML 
is asymmetric because the rectification is
one-sided for $\lambda < 1$, but the reweighting 
steps make the curve more symmetric again.

\begin{figure}[!ht]
\begin{centering}
\includegraphics[width = 0.83\textwidth]
   {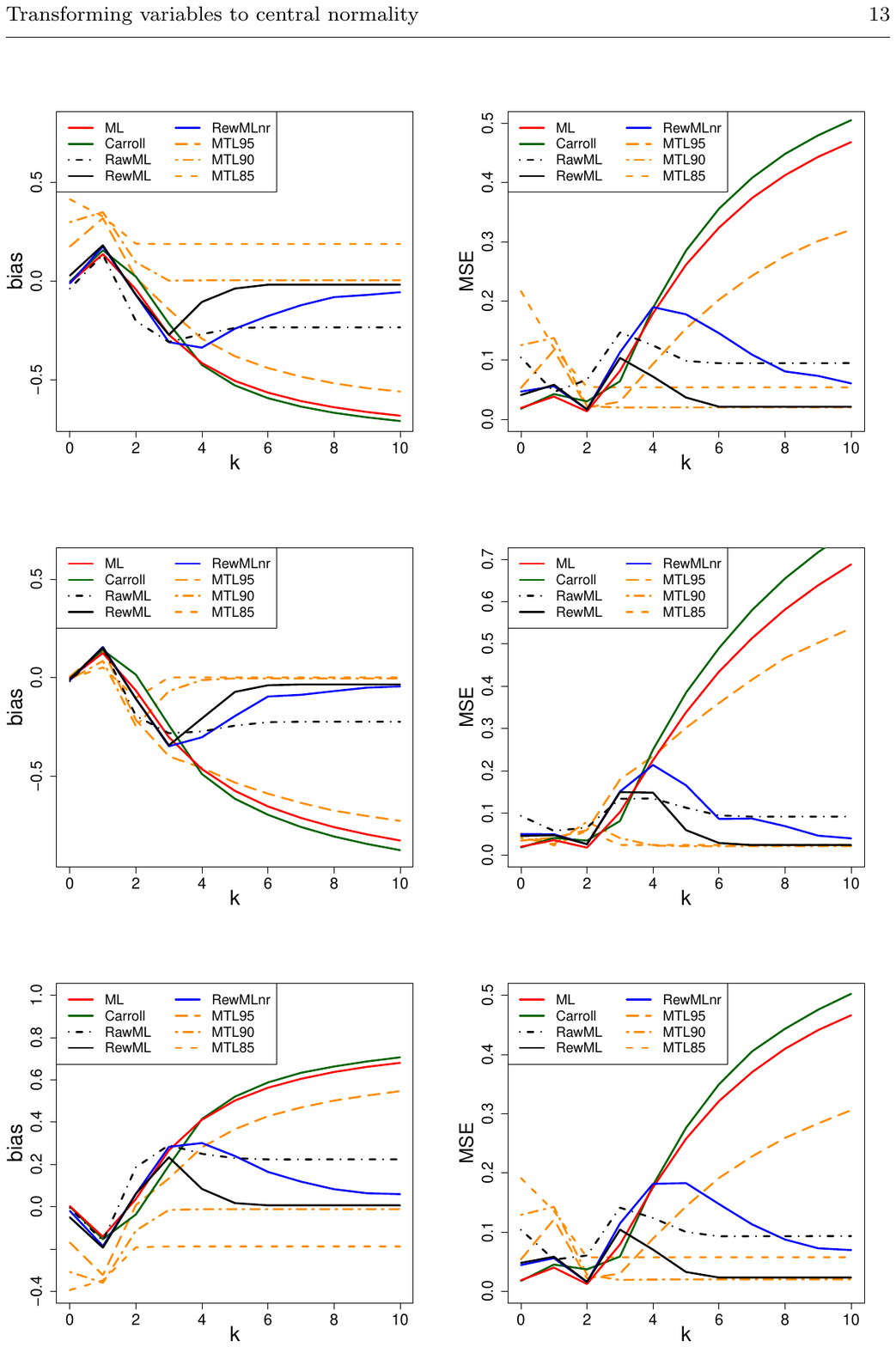}\\
\end{centering}	
\caption{Bias (left) and MSE (right) of the
  estimated $\lambdahat$ of the
  Yeo-Johnson transformation as a function of $k$ 
	which determines how far the outliers are.
	Here the percentage of outliers is fixed at $10\%$\,. 
	The true parameter $\lambda$ used to generate
	the data is 0.5 in the top row, 1.0 in the
	middle row, and 1.5 in the bottom row.}
\label{fig:allkplotsYJ_wraw}
\end{figure}

We now extend Figures \ref{fig:allkplotsYJ}
and \ref{fig:allkplotsBC} of the main text,
which showed the bias and MSE of the estimators 
for 10\% of contamination positioned at an 
increasing deviation $k$ from the center of the 
distribution.
Figure \ref{fig:allkplotsYJ_wraw} shows the 
results for the YJ transformation with the 
initial estimator added. 
We see that the initial estimator behaves roughly
like the reweighted estimator for small $k$,
but the bias does not redescend as the outliers 
are moved further away from the center. 
As a result, its MSE only stabilizes at a 
higher value of $k$.\par
\begin{figure}[!ht]
\begin{centering}
\includegraphics[width = 0.83\textwidth]
   {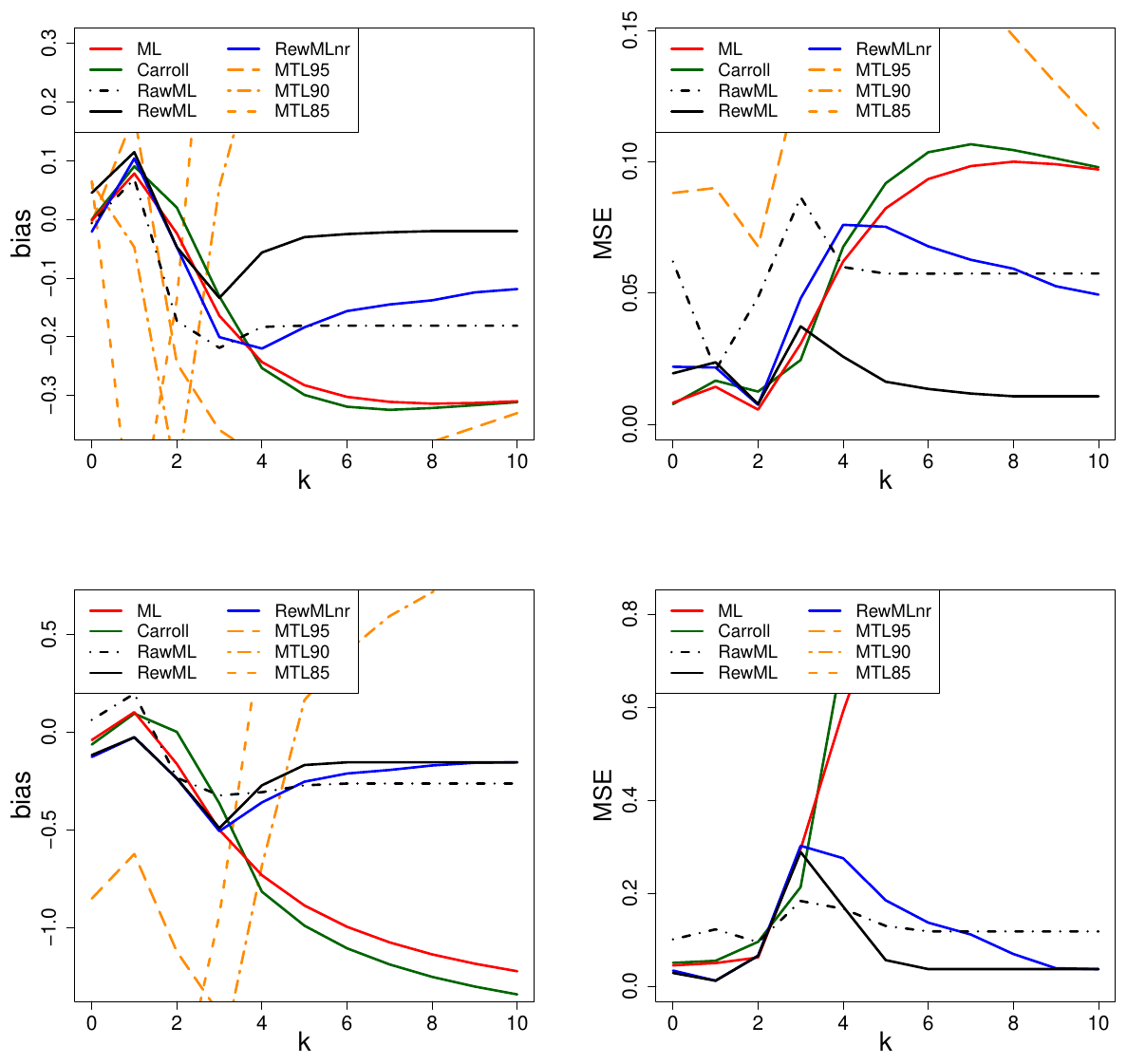}\\
\end{centering}	
\caption{Bias (left) and MSE (right) of the
  estimated $\lambdahat$ of the 
  Box-Cox transformation as a function of $k$ which
	determines how far the outliers are.
	Here the percentage of outliers is fixed at $10\%$\,.
	The true $\lambda$ used to generate	the
	data is 0 in the top row and 1 in the bottom row.}
\label{fig:allkplotsBC_wraw}
\end{figure}

 
Figure \ref{fig:allkplotsBC_wraw} shows the
corresponding results for the BC transformation. 
A similar conclusion can be drawn here.
For both the YJ and BC transforms it is clear
that the application of the redescending weight
function \eqref{eq:weights} reduces the effect 
of far outliers on $\lambdahat$.
	
\end{document}